\documentclass[preprint,12pt]{elsarticle}

\usepackage{graphicx}
\usepackage{multirow}
\usepackage{amsmath}
\usepackage{amssymb}
\usepackage{amsfonts}
\usepackage{amsthm}
\usepackage{mathrsfs}
\usepackage[title]{appendix}
\usepackage{xcolor}
\usepackage{textcomp}
\usepackage{manyfoot}
\usepackage{booktabs}
\usepackage{algorithm}
\usepackage{algorithmicx}
\usepackage{algpseudocode}
\usepackage{listings}
\usepackage{upgreek}
\usepackage{lineno}
\usepackage{longtable}
\usepackage{adjustbox}
\usepackage{tabularx}
\usepackage{hyperref}
\usepackage{threeparttable}
\usepackage{subfig}
\journal{Artificial Intelligence in Medicine}

\begin{document}

\begin{frontmatter}



\title{NPCNet: Navigator-Driven Pseudo Text for Deep Clustering of Early Sepsis Phenotyping}


\author[1]{Pi-Ju Tsai}

\author[2]{Charkkri Limbud}

\author[3,4]{Kuan-Fu Chen\corref{cor1}}
\ead{drkfchen@gmail.com}

\author[5,6]{Yi-Ju Tseng\corref{cor1}}
\ead{yjtseng@nycu.edu.tw}

\cortext[cor1]{Corresponding author}

\affiliation[1]{organization={Institute of Data Science and Engineering, National Yang Ming Chiao Tung University},
    city={Hsinchu},
    country={Taiwan}}
\affiliation[2]{organization={EECS International Graduate Program, National Yang Ming Chiao Tung University},
    city={Hsinchu},
    country={Taiwan}}
\affiliation[3]{organization={College of Intelligent Computing, Chang Gung University},
    city={Taoyuan},
    country={Taiwan}}
\affiliation[4]{organization={Department of Emergency Medicine, Chang Gung Memorial Hospital},
    city={Hsinchu},
    country={Taiwan}}
\affiliation[5]{organization={Department of Computer Science, National Yang Ming Chiao Tung University},
    city={Hsinchu},
    country={Taiwan}}
\affiliation[6]{organization={Computational Health Informatics Program, Boston Children’s Hospital},
    city={Boston},
    country={USA}}

\begin{abstract}
Sepsis is a heterogeneous syndrome. Identifying clinically distinct phenotypes may enable more precise treatment strategies. In recent years, many researchers have applied clustering algorithms to sepsis patients. However, many existing approaches require aggregating or imputing temporal Electronic Health Records (EHRs), which can limit the model's ability to capture disease progression. Furthermore, existing clustering processes seldom incorporate domain-driven constraints, often resulting in phenotypes that lack clear clinical distinction. 
We propose a novel clustering network, NPCNet, that comprises a text embedding generator, a clustering operator, and a target navigator. We first transformed EHRs into pseudo texts by discretizing continuous clinical measurements and integrated them with static variables to construct the embeddings. The target navigator then infused clinical knowledge into the embeddings through auxiliary tasks, constraining clustering results to better align sepsis phenotypes with clinical significance. Finally, the clustering operator employed an iterative refinement mechanism to jointly optimize phenotype centroids and patient representations under domain-driven constraints.
Extensive experiments on public datasets demonstrated that NPCNet achieved superior performance on both internal clustering and clinical metrics. We identify four distinct sepsis phenotypes ($\upalpha$, $\upbeta$, $\upgamma$, and $\updelta$) with divergent SOFA trajectories. Notably, although $\alpha$ and $\delta$ exhibit similarly severe conditions initially, NPCNet distinguishes patients likely to improve ($\alpha$) from those at risk of deterioration ($\delta$). Furthermore, treatment effect analysis suggests that early vasopressor administration may be associated with improved outcomes in the $\alpha$, $\beta$, and $\delta$ phenotypes. These findings highlight NPCNet's potential for phenotype-specific treatment strategies.
\end{abstract}


\begin{highlights}
\item Navigator-guided deep clustering framework for clinically meaningful sepsis phenotyping.
\item Chronological pseudo-text encoding for preserving irregular EHR patterns.
\item Four distinct sepsis phenotypes with divergent SOFA trajectories are identified.
\item Early vasopressor administration yields distinct phenotype-level outcomes.
\end{highlights}

\begin{keyword}
Sepsis Phenotyping \sep Deep Clustering \sep Deep Learning \sep Electronic Health Records \sep Pseudo Text


\end{keyword}

\end{frontmatter}



\section{Introduction}
Sepsis is a life-threatening condition that occurs when the body’s immune system has an extreme response to an infection that causes organ dysfunction \cite{singer2016third}. An estimated 11 million deaths worldwide were associated with sepsis, accounting for approximately 20\% of all global deaths, while survivors often experience poor long-term outcomes \cite{rudd2020global}\cite{prescott2019understanding}. Activation of timely interventions is critical to improving clinical outcomes \cite{prescott2017improving}\cite{evans2021surviving}. For patients with sepsis, each hour of treatment delay has been associated with a 4–8\% increase in risk of death \cite{seymour2017time}. 

However, the progression of sepsis varies widely depending on host factors, infection etiology, or the involvement of multiple organ dysfunctions. The inherent heterogeneity dictates that a uniform therapeutic strategy is unlikely to be effective for all patients \cite{giamarellos2024pathophysiology}. Consequently, providing patient-specific care by synthesizing information from multiple sources becomes imperative, yet the primary barrier to this approach is the difficulty of identifying clinically meaningful patterns within complex and temporal electronic health records (EHRs). 

To characterize clinical heterogeneity in sepsis, previous studies have employed clustering algorithms to explore distinct sepsis phenotypes \cite{seymour2019derivation}\cite{xu2022sepsis}\cite{yin2020identifying}. These studies suggest that patient stratification may facilitate more personalized treatment strategies \cite{giamarellos2024pathophysiology}. However, clinical measurements are collected at irregular, clinical driven frequencies, with vital signs typically recorded more frequently than laboratory tests. Most existing approaches address this irregularity through feature aggregation \cite{guo2020evaluation} or imputation \cite{groenwold2020informative}, potentially obscuring clinically informative temporal dynamics, such as measurement frequency \cite{xie2022deep} and value fluctuations, which are often clinical indicators of physiological deterioration.

Another important challenge is ensuring that the resulting phenotypes are clinically meaningful. Most existing clustering methods, ranging from statistical methods to deep clustering architectures, primarily focus on optimizing mathematical objectives, such as maximizing intra-cluster compactness or inter-cluster separation. Although these models can identify patients who are statistically distinct in a high-dimensional latent space, they often lack mechanisms to incorporate domain-specific knowledge that reveals clinical significance, such as important patient outcomes, into the learning process \cite{zbMATH03129892}\cite{xie2016unsupervised}. Consequently, clusters with favorable mathematical properties may not necessarily represent clinically meaningful phenotypes or provide actionable insights for real-world medical decision-making.

To address these limitations, we propose a novel deep clustering framework, NPCNet, to identify clinically meaningful sepsis phenotypes from temporal EHRs. NPCNet uses a text embedding generator that transforms patient EHRs into representations. Specifically, for the temporal variables, we sequence the clinical events into pseudo texts to preserve their inherent chronological occurrence and employ a discretization strategy to quantize continuous clinical measurements into bins. We then encode static variables into an embedding space and inject them into pseudo texts. This mechanism integrates multi-modal patterns without sacrificing their unique distributions. Furthermore, NPCNet incorporates a target navigator to guide the embeddings for clinical relevance through two auxiliary tasks. One of the tasks infuses clinical knowledge by directly encouraging embeddings to reflect the discharge status of patients. The other captures relative relationships among patients in the embedding space. By jointly optimizing these objectives, NPCNet effectively bridges the gap between clustering performance and clinical relevance.

The main contributions of this work are as follows:
\begin{itemize}
    \item We propose NPCNet, a novel deep clustering framework that integrates temporal EHR representation with clinical guidance. Specifically, we develop a pseudo-text representation strategy that preserves inherent temporal information in EHRs without aggregation or imputation, and introduce a target navigator to incorporate clinical knowledge into the clustering process.
    \item We evaluate NPCNet using the MIMIC-IV dataset and externally validate the generalizability on the eICU dataset. NPCNet achieves superior performance over most baseline clustering models across multiple internal metrics and a trajectory-based clinical metrics. We also introduce the Trajectory Divergence Index (TDI) to quantify the clinical distinction between identified phenotypes based on their SOFA trajectories.
    \item We identify four clinically distinct sepsis phenotypes ($\upalpha$, $\upbeta$, $\upgamma$, and $\updelta$) with divergent SOFA trajectories. Notably, NPCNet distinguishes $\upalpha$ patients, who tend to improve, from $\updelta$ patients, who are at risk of deterioration, despite their similarly severe initial conditions. Furthermore, treatment effect analysis reveals phenotype-specific associations between early vasopressor administration and patient outcomes, suggesting potential opportunities for phenotype-guided treatment strategies.
\end{itemize}

\section{Materials and Methods}
\subsection{Datasets}
We used the Medical Information Mart for Intensive Care (MIMIC)-IV version 2.2 database as the main dataset \cite{johnson2023mimic}, which contains patients who have ever been admitted to the emergency department or an intensive care unit at Beth Israel Deaconess Medical Center (BIDMC) in Boston, MA, from 2008 to 2022. Sepsis episodes were identified according to Sepsis-3 criteria \cite{shankar2016developing} during the first ICU stay of each hospitalization. After applying the eligibility criteria (\textbf{Supplementary Figs. 1} and  \textbf{2}), 19,834 episodes were included for clustering analysis. A further 2,013 episodes met the criteria for treatment effect analysis (\textbf{Supplementary Fig. 3}).

For external validation, we further utilized the eICU Collaborative Research Database \cite{pollard2018eicu}, a multi-center intensive care database containing de-identified health records from over 200 hospitals across the United States between 2014 and 2015.
Sepsis cases in the eICU dataset were identified based on publicly available code from prior work \cite{moor2023predicting}. For treatment effect analysis, the same selection criteria as those applied to the MIMIC-IV dataset were used. Finally, we identified 13,660 cases for clustering analysis (\textbf{Supplementary Fig. 4}) and 1,235 cases for treatment effect analysis (\textbf{Supplementary Fig. 5}).

We included 65 variables for clustering \cite{seymour2019derivation}, including two demographics, 31 Elixhauser comorbidities, 28 laboratory test results, and four vital signs. Elixhauser comorbidities index were derived from diagnosis codes recorded during prior hospitalizations or emergency department visits while other variables were extracted from the first six hours after ICU admission. The details of the variables are presented in \textbf{Supplementary Table 2} for the development dataset and \textbf{Supplementary Table 3} for the external validation dataset. Variable ranges were defined based on prior studies \cite{seymour2019derivation}\cite{moor2023predicting}, and clinically invalid or placeholder values (e.g., 999999) were excluded.

\begin{figure*}[!htbp]
    \centering
    \includegraphics[width=\textwidth]{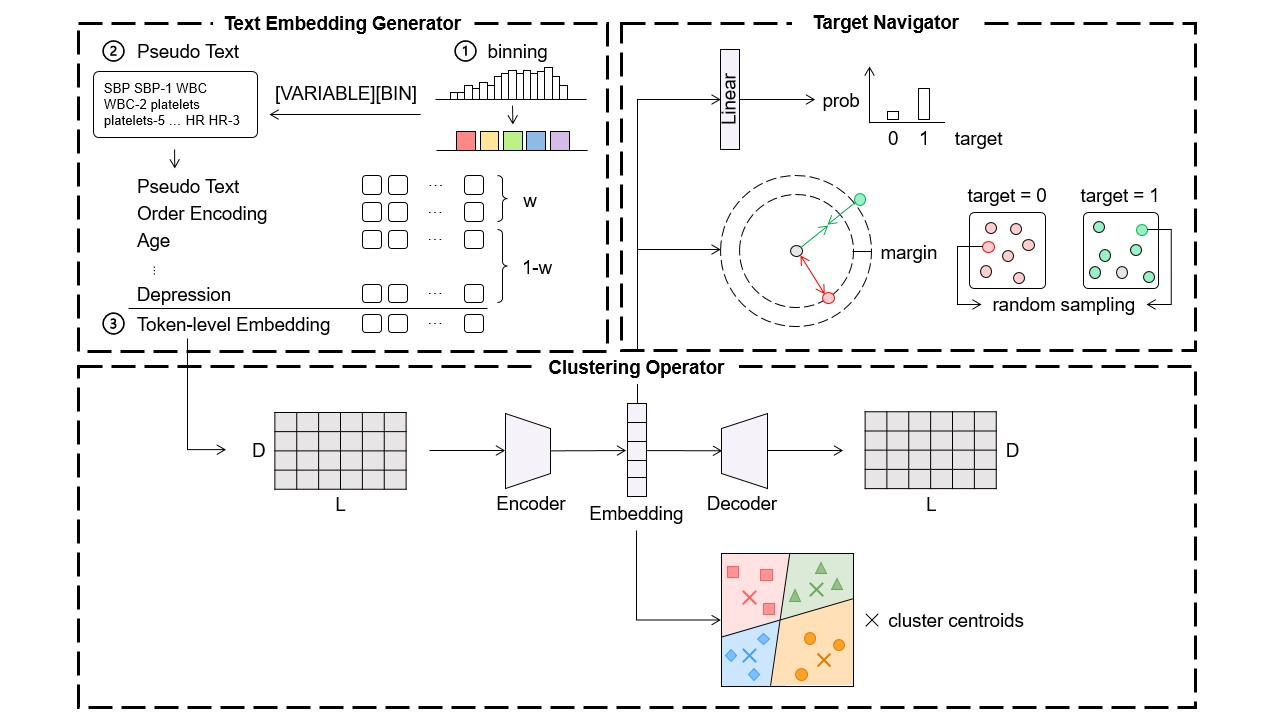}
    \caption{
    The overview of NPCNet. Through the text embedding generator, we first bin the value of time-varying variables into bin indices according to the distribution of the training set. We then transform the series of time-varying variables into pseudo texts, combining static information. NPCNet trains with an objective function that consists of $\mathcal{L_\text{rec}}$, $\mathcal{L_\text{clustering}}$, and $\mathcal{L_\text{navigator}}$. Through the clustering network in the clustering operator and clinical relevance through the target navigator, we can derive the patient embeddings and the centroids of computable phenotypes during the training stage.
    }
    \label{fig1}
\end{figure*}

\subsection{Model Architecture}
\textbf{Fig.} \ref{fig1} illustrates the overall architecture of NPCNet, which consists of a text embedding generator, a target navigator, and a clustering operator.

\subsection{Text Embedding Generator}
\begin{figure}[!htbp]
    \centering
    \includegraphics[width=\textwidth]{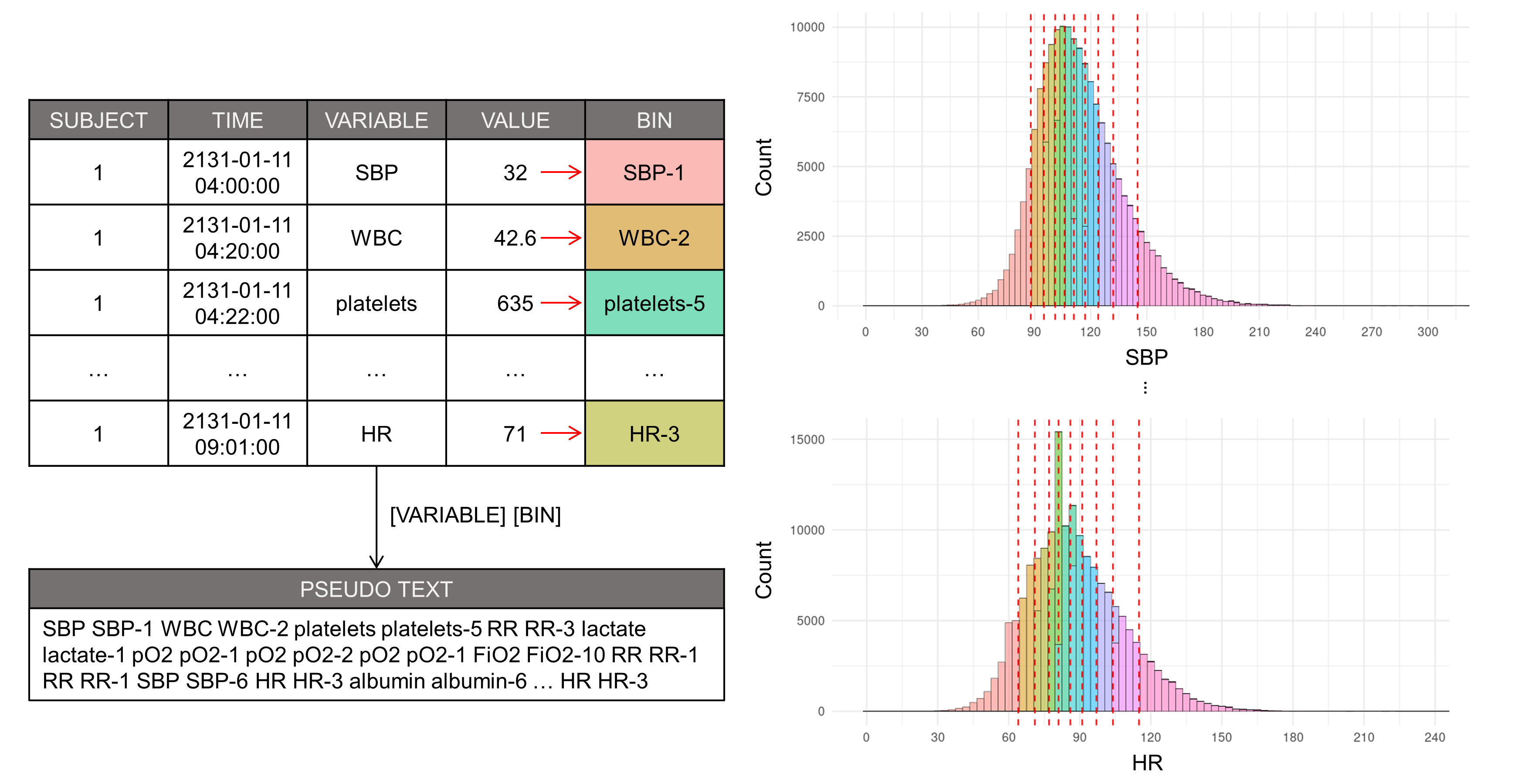}
    \caption{Illustration of the binning process for converting time-varying EHR into a pseudo-text sequence. Numerical observations are discretized into quantile-based bins, encoded using the [VARIABLE][BIN] format, and arranged chronologically to generate the final pseudo text. For example, a systolic blood pressure (SBP) value of 32 falls into the lowest quantile bin (SBP-1) and is encoded as "SBP SBP-1".}
    \label{fig2}
\end{figure}

We extracted static and time-varying variables separately and then integrated them to form a token-level embedding (\textbf{Fig.} \ref{fig1}, upper left). For time-varying variables, we collected vital signs and laboratory tests within the first six hours after ICU admission. Because the tokenization-embedding paradigm does not work well with numerical values \cite{lin2020birds}, we first discretized each variable into $B$ qualitative bins based on the quantiles of the training set. This binning transformed numerical observations into discrete representations while preserving their relative ranges. Each observation was then represented using the format [VARIABLE][BIN], and all resulting tokens were arranged chronologically according to the examination time to form a pseudo-text sequence of length $l$ \cite{hegselmann2023tabllm}. An example of this conversion process is illustrated in \textbf{Fig.} \ref{fig2}.

To convert pseudo text into token-level embeddings, we first split the pseudo text by spaces into a sequence of $l$ tokens, where $l$ denotes the sequence length. Each token was mapped to an index in the vocabulary $\mathcal{V}$, resulting in the token-index sequence  $[t_0, t_1, \dots, t_{l-1}]$, where $t_i \in \mathcal{V}$. We then obtained the token-level embedding matrix $P \in \mathbb{R}^{l \times d}$ from the embedding matrix $v \in \mathbb{R}^{|\mathcal{V}| \times d}$ as:
$$P = [e_0, e_1, \dots, e_{l-1}],$$
where $e_i = v[t_i]$ and $d$ is the embedding dimension.

To incorporate the temporal structure of the time-varying variables, we introduced an order encoding, which was similar to the positional encoding in Transformer \cite{vaswani2017attention}, but corresponded to the examination order of variables here. Let $O \in \mathbb{R}^{l \times d}$ denote the matrix stacking the order encodings OE of all positions:
$$O = [\text{OE}_{(0)}, \text{OE}_{(1)}, \dots, \text{OE}_{(l-1)}],$$

Static variables, such as demographics and comorbidities, were encoded separately \cite{rupp2023exbehrt}. Each category $c_i$ of the $i$-th variable was represented by an embedding from $W_i$, and the embeddings were summed to derive the static embedding $S \in \mathbb{R}^{d}$:
$$ S = \sum_{i}W_i[c_i]. $$

\begin{figure}[!htbp]
    \centering
    \includegraphics[width=\textwidth]{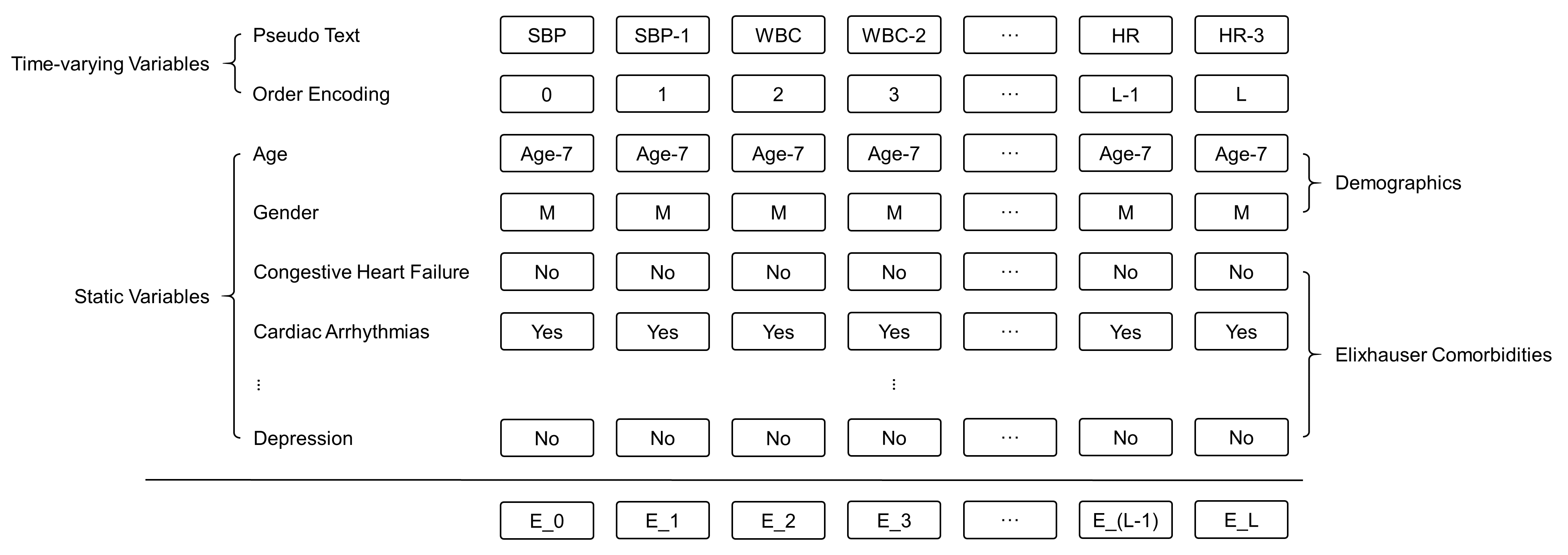}
    \caption{An example of the NPCNet input representation. Temporal clinical variables are converted into pseudo text and mapped to token-level embeddings, which are combined with order encoding to preserve the examination order. Static variables are encoded separately and fused with the temporal representation using a weighted sum to form the final input for NPCNet.}
    \label{fig3}
\end{figure}

Finally, we combined (1) the pseudo text embedding $P$ with order encoding $O$, and (2) the static embedding $S$ using different weights, resulting in the input $x \in \mathbb{R}^{l \times d}$:
$$x = w \times (P + O) + (1-w) \times S,$$
where $w \in [0, 1]$ is a hyperparameter that controls the contribution of static and time-varying variables. The example is shown in \textbf{Fig.} \ref{fig3}.


\subsection{Target Navigator}
To enhance the clinical relevance of clustering results, we introduced the target navigator. Through applying two auxiliary tasks, the navigator enabled the embedding $E$ to progressively learn clinical relevance through backpropagation while simultaneously refining the assignment of phenotypes. When the embedding $E$ contained more clinical relevance, the clustering results can uncover more heterogeneity among phenotypes.

Taking the discharge status navigator as an example, the target navigator guided the learning of the embedding $E$ through two auxiliary tasks. First, it encouraged embedding $E$ to capture whether a patient was alive or not at discharge \cite{naviDCN}. Second, it incorporated relational knowledge by constraining the relative positions of patients in the embedding space. Patients with the same discharge status were encouraged to be closer while keeping patients with different discharge statuses away from each other.
The embedding $E$ was passed through a linear layer, followed by a softmax activation to output the probabilities for each discharge status.
\begin{equation}
    p=\mathrm{softmax}(W^T E+b),
\end{equation}
where $W \in \mathbb{R}^{d \times c}$ and $b \in \mathbb{R}^{c}$ are learnable parameters, and $c$ is the number of categories.
By comparing these probabilities with the actual clinical status $y$, the probability loss is as follows:
\begin{equation}
\begin{aligned}
    \mathcal{L}_{\text{prob}} = - \frac{1}{N} \sum_{j=1}^{N} \sum_{i=1}^{c} w_i (1 - p_{t,i}^j)^\upgamma \log(p_{t,i}^j), \\
    p_{t,i}^j =
    \begin{cases} 
    p_i^j & \text{if } y^j = i \\
    1 - p_i^j & \text{otherwise},
    \end{cases}
\end{aligned}
\end{equation}
where $N$ is the number of patients, $w_i$ is the weight hyperparameter, $p_i^j$ is the probability of discharge status $i$ for patient $j$, and $\upgamma$ is the focusing hyperparameter. $p_{t,i}^j$ denotes the probability corresponding to the target label for class $i$, determined by $y^j$.
$\upgamma$ controls the contribution of each patient. For patients whose probabilities for each status are close to each other, the contribution to $\mathcal{L}_{\text{prob}}$ is larger. The value is lower for the patient with a high probability of a certain discharge status, thus encouraging the model to focus more on challenging ones.

Then, the model adopted a triplet strategy. For each anchor ($a$), we randomly sampled a positive sample ($p$) with the same discharge status as the anchor, and a negative sample ($n$) with a different discharge status, both drawn from the training set. The triplet loss encourages the embedding of the anchor to be closer to the positive patient than to the negative patient by at least a predefined margin. Specifically, the distance loss is defined as:
\begin{equation} 
    \mathcal{L}_{\text{dist}} = \text{max(d($a$, $p$) - d($a$, $n$) + margin, 0)}
\end{equation}

where d($\cdot, \cdot$) is the Euclidean distance, and $\text{margin}$ is a hyperparameter specifying the minimum desired separation between the positive and negative samples. This objective encourages patients with the same discharge status to have closer embeddings while separating patients with different discharge statuses, thereby incorporating clinically relevant information into the learned embedding space.

The overall target navigator loss was the sum of two losses with different weights:
\begin{equation}
    \mathcal{L}_{\text{navigator}} = \kappa_1 * \mathcal{L}_{\text{prob}} + \kappa_2 * \mathcal{L}_{\text{dist}},
\end{equation}
where $\kappa_1$ and $\kappa_2$ are hyperparameters.

\subsection{Clustering Operator}
We used Deep Clustering Network (DCN) \cite{yang2017towards} as the clustering operator (\textbf{Fig.} \ref{fig1}, bottom). DCN adopts an alternating stochastic optimization approach that jointly optimizes reconstruction loss $\mathcal{L_\text{rec}}$ and clustering loss $\mathcal{L_\text{clustering}}$. 
 
\begin{equation}
    \mathcal{L_\text{rec}} = \sum^N_{i=1}\parallel x_i - \hat{x}_i \parallel^2_2,
\end{equation}
where $N$ is the number of patients and $\hat{x}$ is the reconstruction of $x$ from the output of the decoder.
\begin{equation}
\begin{aligned}
    \mathcal{L_\text{clustering}} = \sum^N_{i=1}\parallel E_i - Ms_i \parallel^2_2\\
    \mathrm{\text{s.t.}}\ s_{j, i} \in \{0, 1\}, \mathbf{1}^T s_i=1,\ \forall i, j,
\end{aligned}
\end{equation}
where $M \in \mathbb{R}^{d \times k}$ is the matrix including cluster centroids, $k$ is the number of clusters, and $s_i$ is the one-hot assignment vector.

\subsection{Objective Function}
Ultimately, the overall loss function of NPCNet was as follows:
\begin{equation}
    \mathcal{L} = \lambda_1 * \mathcal{L}_{\text{rec}} + \lambda_2 * \mathcal{L}_{\text{clustering}} + \lambda_3 * \mathcal{L}_{\text{navigator}},
\end{equation}
where $\lambda_1$, $\lambda_2$, and $\lambda_3$ are hyperparameters to balance three components. 
Then, we leveraged $\mathcal{L}$ as the overall objective to adjust the network weights during backpropagation.

\subsection{Clinical Significance Analysis}
To assess the clinical relevance of phenotypes, we compared short-term SOFA trajectories across phenotypes. For each patient, hourly SOFA scores were calculated from 7 to 24 hours after ICU admission \cite{vincent1996sofa}. Patients were stratified by their SOFA scores at the sixth hour after ICU admission into six severity groups: [0, 1], [2, 3], [4, 5], [6, 7], [8, 9], and [10, 24] \cite{ferreira2001serial}, reducing the confounding effect of patients' initial severity.

Within each stratum, phenotype pairs were compared using a Generalized Additive Mixed Model (GAMM) to account for nonlinear temporal patterns and the variability across patients.
Next, we introduced the Trajectory Divergence Index (TDI) according to GAMM to quantify clinical relevance.
TDI is the proportion of statistically significant pairwise differences in SOFA trajectories within each stratum and across all time points:
\begin{equation}
\text{TDI} = \frac{\text{the number of statistically significant pairs}}{\text{the number of pairwise comparisons}}
\end{equation} 
For example, with four phenotypes, there are six cluster pairs per stratum, 18 time points, resulting in a total of 648 pairwise comparisons. TDI ranges from 0 to 1, with higher values indicating greater divergence in short-term SOFA trajectories among phenotypes.

\subsection{Treatment Effect Analysis}
To assess whether phenotype modifies treatment-outcome associations, we employed multivariable logistic regression with in-hospital mortality as the outcome. The treatment variables were (1) the volume of IV fluids, defined as the cumulative fluid input within 12 hours after initial hypotension (MAP $<$ 65), and (2) the time to vasopressor, defined as the duration from initial hypotension to vasopressor initiation. \textbf{Supplementary Table 1} provides descriptions of the variables in the logistic regression. Statistical significance of the treatment and interaction terms was assessed using the Wald test.



\subsection{Implementation Details}
To prevent data leakage due to the target navigator, the MIMIC-IV dataset was split by patients, as individual patients could contribute multiple episodes. The MIMIC-IV dataset was divided into a training set (80\%) and a testing set (20\%). Thresholds for the binning task were derived exclusively from the training set and subsequently applied to the testing set.
During the training stage, the target navigator was incorporated to guide the model to learn clinical knowledge, whereas in the testing stage, the navigator was removed, and model inputs were restricted to information from the first six hours after ICU admission.
The eICU dataset served as the external validation dataset. To assess the model's generalizability and robustness across different datasets, the binning thresholds and the model derived from the MIMIC-IV training set were directly applied to the eICU dataset without re-training.

We explored the number of phenotypes ($k$) using multiple internal clustering metrics across different combinations of cluster numbers, distance measures, and clustering methods. However, these evaluations yielded inconclusive results, with no consistent preference for a specific $k$. We therefore set k to four, consistent with previous sepsis phenotyping studies \cite{seymour2019derivation}\cite{xu2022sepsis}\cite{zhang2018identification}, to ensure comparability, and employed K-means to initialize the cluster centroids. Hyperparameters were selected by random search, with model settings provided in \textbf{Supplementary Table 4}. The training and testing sets showed no statistically significant differences in almost all variables (\textbf{Supplementary Table 5}). Each experiment was run 10 times with different random seeds, and all benchmark and ablation comparisons used the same testing set to maintain consistency across evaluations.

\subsection{Performance Metrics}
Since sepsis phenotypes lack a universal definition, we focused on internal metrics, which objectively assess the clustering results. Internal metrics focus on the compactness within a cluster and the separation between clusters. We chose three common internal metrics, the Silhouette Index (SI), the Calinski-Harabasz Index (CHI), and the Davies-Bouldin Index (DBI). Additionally, to assess clinical relevance, we designed a clinical metric, TDI, that demonstrates the model’s ability to produce phenotypes that can identify short-term organ failure trajectories.

\subsection{Statistical analysis}
Continuous variables were summarized as mean (SD) or median (IQR) according to their distribution, and categorical variables as number (\%). Between-group comparisons used the t-test or Mann-Whitney U test for continuous variables and the chi-square test for categorical variables. Differences between clustering models were assessed using a two-sample t-test. The significance level was set to 0.05 for statistical significance.

\section{Results}
\subsection{Cohort characteristics of sepsis phenotypes}
\begin{figure*}[!htbp]
    \centering
    \includegraphics[width=\textwidth]{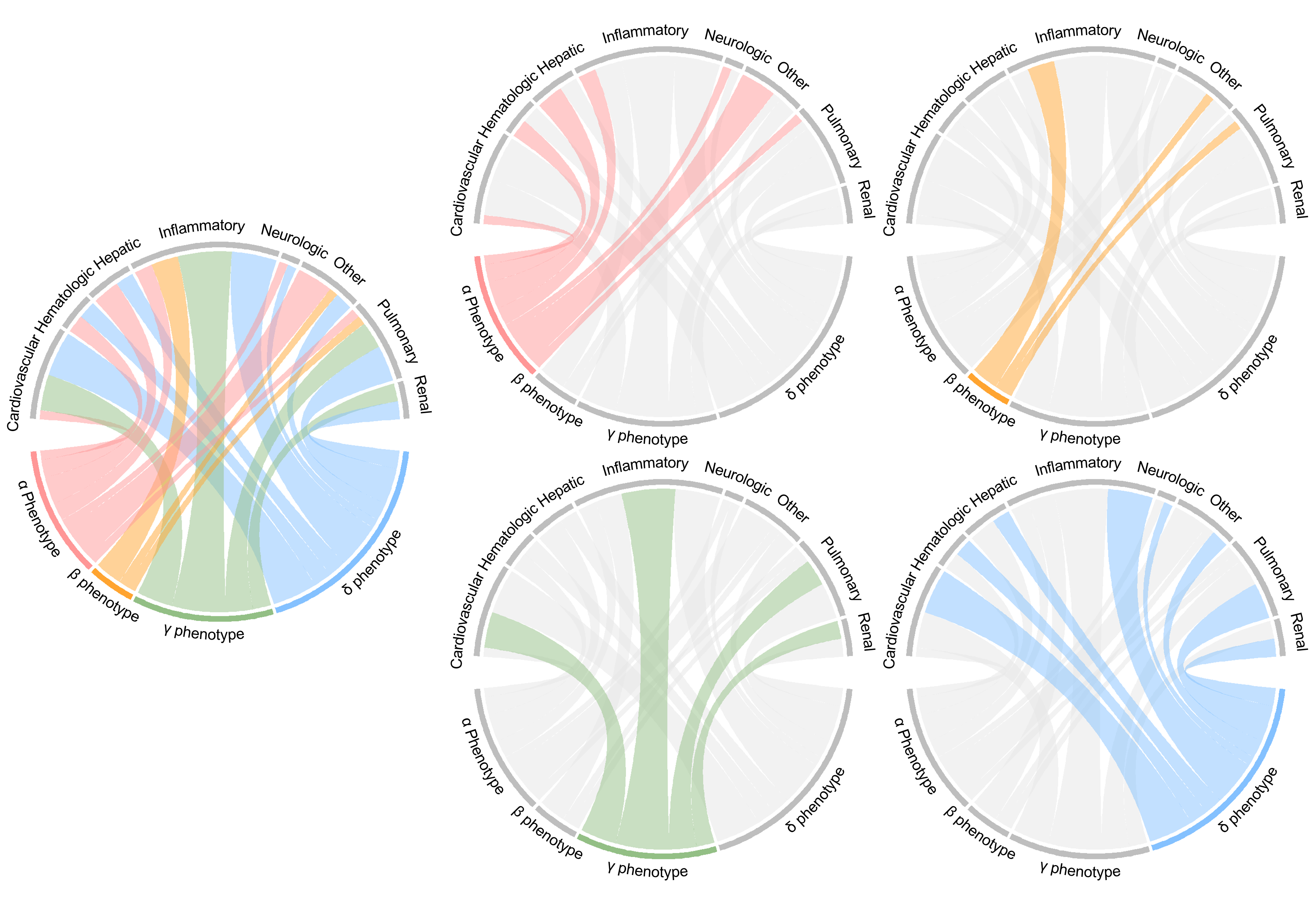}
    \caption{
    Abnormal clinical variables, grouped into eight organ systems \cite{seymour2019derivation}, among the sepsis computable phenotypes. The ribbon connects from a phenotype to an organ system if the group median is more abnormal than the overall median (\textbf{Supplementary Table 2}). The more clinical variables are abnormal for the phenotype, the broader the ribbon.}
    \label{fig4}
\end{figure*}

Clustering identified four computable phenotypes of $\upalpha$, $\upbeta$, $\upgamma$, and $\updelta$ with their clinical characteristics summarized in \textbf{Table} \ref{table1} and \textbf{Supplementary Table 6}. 
The $\upalpha$ phenotype was younger with few comorbidities, while the $\updelta$ phenotype was likely older with more comorbidities (\textit{p}-values $< 0.001$). 
Despite these differences, both phenotypes demonstrated abnormalities in almost all organ systems, especially in hepatic and hematologic systems, indicating severe conditions at the time of early ICU admission. These observations were based on laboratory measurements relative to the overall medians (\textbf{Fig.} \ref{fig4} \cite{gu2014circlize}\cite{seymour2019derivation}).
The $\upbeta$ phenotype exhibited abnormalities mainly in the inflammatory and other systems, with few abnormal variables in the remaining organ systems whereas the $\upgamma$ phenotype was more likely to have elevated inflammatory markers (such as CRP and neutrophils) and renal dysfunction.

Despite similar multi-organ abnormalities (\textbf{Fig.} \ref{fig4}), $\upalpha$ and $\updelta$ phenotypes showed markedly different outcomes (\textbf{Table} \ref{table1}). The $\upalpha$ phenotype had the lowest in-hospital mortality rate at 1.4\%, whereas the $\updelta$ phenotype showed the highest at 46.2\%. Among the $\updelta$ phenotype, approximately 70.4\% died either in hospital or within 365 days after discharge.

\begin{table*}[htbp]
\caption{Characteristics of the 4 sepsis computable phenotypes.}
\label{table1}
\resizebox{\textwidth}{!}{
\begin{tabular}{llllll}
\toprule
\textbf{Characteristic} & \textbf{Total} & $\boldsymbol{\upalpha}$ & $\boldsymbol{\upbeta}$ & $\boldsymbol{\upgamma}$ & $\boldsymbol{\updelta}$ \\
\midrule
No. of patients (\%)                   & 19834 (100) & 4624 (23.3) & 6246 (31.5) & 6250 (31.5) & 2714 (13.7) \\
Age, mean (SD)                         & 64.7 (15.8) & 63.3 (14.2) & 63.6 (16.2) & 66.2 (16.1) & 66.1 (16.3) \\
Sex, No. (\%) \\
\hspace{1em}Male                       & 11763 (59.3) & 3122 (67.5) & 3535 (56.6) & 3513 (56.2) & 1593 (58.7) \\
\hspace{1em}Female                     & 8071 (40.7) & 1502 (32.5) & 2711 (43.4) & 2737 (43.8) & 1121 (41.3) \\
SOFA score, median (IQR)               & 7 [4, 10] & 7 [3, 9] & 5 [3, 8] & 7 [5, 10] & 11 [8, 14] \\
Elixhauser comorbidities, median (IQR) & 0 [0, 5] & 0 [0, 0] & 0 [0, 5] & 4 [0, 6] & 5 [0, 7] \\
\midrule
\textbf{Inflammation} \\
Temperature, mean (SD)                 & 37.0 (0.9) & 36.8 (0.7) & 37.1 (0.8) & 37.0 (0.8) & 36.9 (1.1) \\
Bands, median (IQR)                    & 3.0 [1.4, 5.6] & 2.4 [1.2, 4.4] & 3.0 [1.4, 5.4] & 3.2 [1.6, 6.0] & 4.0 [2.0, 7.4] \\
CRP, median (IQR)                      & 81 [47, 121] & 53 [32, 83] & 81 [47, 122] & 95 [60, 132] & 104 [72, 137] \\
ESR, median (IQR)                      & 49.0 [36.0, 63.6] & 48.8 [37.0, 62.4] & 49.0 [35.7, 63.8] & 49.2 [35.8, 63.8] & 49.0 [35.6, 64.0] \\
WBC, median (IQR)                      & 12.8 [9.2, 17.3] & 13.0 [9.7, 16.7] & 12.1 [8.7, 16.3] & 12.9 [9.1, 17.6] & 14.4 [9.9, 20.0] \\
Neutrophils, median (IQR)               & 81.3 [75.0, 86.4] & 79.3 [74.2, 83.5] & 81.3 [74.7, 86.3] & 82.5 [75.9, 87.8] & 83.0 [75.8, 88.2] \\
Metamyelocytes, median (IQR)           & 0.4 [0.2, 1.0] & 0.4 [0.2, 0.8] & 0.4 [0.2, 1.0] & 0.4 [0.0, 1.0] & 0.8 [0.2, 1.6] \\
Myelocytes, median (IQR)               & 0.2 [0.0, 0.4] & 0.2 [0.0, 0.4] & 0.2 [0.0, 0.4] & 0.0 [0.0, 0.4] & 0.2 [0.0, 0.6] \\
Promyelocytes, median (IQR)            & 1.2 [1.0, 1.6] & 1.2 [1.0, 1.6] & 1.2 [1.0, 1.6] & 1.2 [1.0, 1.6] & 1.2 [1.0, 1.6] \\
Lymphocytes, median (IQR)              & 11.0 [7.0, 16.4] & 14.5 [10.5, 19.0] & 11.1 [7.2, 16.3] & 9.3 [5.6, 14.0] & 8.7 [5.0, 13.7] \\
\midrule
\textbf{Pulmonary} \\
SaO2, median (IQR)                     & 84.2 [73.8, 93.0] & 86.2 [76.8, 94.6] & 84.6 [74.7, 92.4] & 83.8 [72.0, 92.4] & 81.0 [67.0, 92.0] \\
PO2, mean (SD)                         & 101 (75) & 142 (80) & 97 (69) & 85 (66) & 78 (68) \\
FiO2, mean (SD)                        & 76.5 (21.9) & 85.2 (20.0) & 73.0 (20.7) & 72.9 (21.9) & 77.9 (23.5) \\
Respiratory rate, mean (SD)            & 25.2 (7.6) & 21.6 (6.6) & 24.8 (7.0) & 26.7 (7.4) & 28.9 (7.9) \\
\midrule
\textbf{Cardiovascular} \\
Bicarbonate, mean (SD)                 & 21.8 (5.3) & 23.1 (3.3) & 22.5 (5.0) & 21.5 (5.7) & 18.4 (6.2) \\
Heart rate, mean (SD)                  & 98 (21) & 91 (15) & 97 (20) & 101 (22) & 106 (24) \\
Lactate, median (IQR)                  & 2.3 [1.6, 3.5] & 2.3 [1.7, 2.9] & 2.1 [1.5, 3.0] & 2.4 [1.6, 3.6] & 3.8 [2.2, 7.3] \\
SBP, median (IQR)                      & 94 [85, 107] & 96 [87, 106] & 97 [87, 110] & 93 [83, 106] & 89 [78, 100] \\
Troponin, median (IQR)                 & 0.2 [0.1, 0.4] & 0.2 [0.1, 0.5] & 0.2 [0.1, 0.4] & 0.2 [0.1, 0.4] & 0.2 [0.1, 0.6] \\
\midrule
\textbf{Renal} \\
BUN, median (IQR)                      & 24 [16, 40] & 17 [13, 22] & 22 [15, 35] & 32 [19, 50] & 36 [23, 56] \\
Creatinine, median (IQR)               & 1.2 [0.8, 2.0] & 0.9 [0.7, 1.1] & 1.1 [0.8, 1.7] & 1.4 [1.0, 2.4] & 1.8 [1.2, 2.8] \\
\midrule
\textbf{Hepatic} \\
AST, median (IQR)                      & 66 [36, 143] & 72 [43, 136] & 60 [33, 128] & 61 [32, 133] & 92 [41, 293] \\
ALT, median (IQR)                      & 42 [24, 92] & 46 [28, 89] & 40 [22, 85] & 39 [21, 85] & 53 [25, 167] \\
Bilirubin, median (IQR)                & 0.9 [0.5, 1.7] & 0.9 [0.6, 1.5] & 0.9 [0.5, 1.8] & 0.8 [0.5, 1.8] & 0.9 [0.5, 1.9] \\
\midrule
\textbf{Hematologic} \\
Hemoglobin, mean (SD)                  & 10.1 (2.4) & 9.5 (2.1) & 10.2 (2.3) & 10.3 (2.4) & 10.3 (2.6) \\
INR, median (IQR)                      & 1.4 [1.2, 1.7] & 1.3 [1.2, 1.5] & 1.3 [1.2, 1.6] & 1.4 [1.2, 1.8] & 1.6 [1.2, 2.4] \\
Platelets, median (IQR)                & 174 [120, 241] & 155 [119, 202] & 182 [126, 253] & 187 [123, 256] & 172 [102, 247] \\
\midrule
\textbf{Neurologic} \\
GCS, mean (SD)                         & 9.2 (4.9) & 7.3 (5.0) & 10.8 (4.6) & 9.9 (4.6) & 7.0 (4.4) \\
\midrule
\textbf{Others} \\
Albumin, mean (SD)                     & 3.2 (0.5) & 3.2 (0.5) & 3.2 (0.5) & 3.2 (0.6) & 3.1 (0.6) \\
Chloride, mean (SD)                    & 105 (7) & 108 (5) & 104 (7) & 104 (8) & 105 (9) \\
Glucose, median (IQR)                  & 154 [121, 200] & 162 [136, 190] & 145 [115, 192] & 149 [116, 202] & 174 [128, 250] \\
Sodium, mean (SD)                      & 139 (6) & 140 (3) & 139 (5) & 139 (7) & 140 (7) \\
\midrule
\textbf{Outcomes} \\
In-hospital mortality, No. (\%)              & 3159 (15.9) & 64 (1.4) & 508 (8.1) & 1332 (21.3) & 1255 (46.2) \\
365-day outpatient mortality, No. (\%)       & 5355 (27.0) & 597 (12.9) & 1918 (30.7) & 2185 (35.0) & 655 (24.1) \\
Length of stay in the ICU, median (IQR)      & 3.0 [1.8, 5.9] & 2.1 [1.3, 3.3] & 2.8 [1.8, 5.3] & 3.8 [2.2, 7.2] & 4.8 [2.6, 9.3] \\
Length of stay in the hospital, median (IQR) & 8.3 [5.2, 14.4] & 6.5 [5.0, 10.0] & 8.5 [5.4, 14.5] & 9.8 [5.9, 16.7] & 9.6 [4.5, 17.0] \\
Day of mechanical ventilation, median (IQR)  & 2.08 [0.93, 4.83] & 1.29 [0.80, 2.67] & 1.83 [0.80, 4.25] & 2.72 [1.21, 6.04] & 3.88 [1.69, 8.26] \\
Volume of IV fluids (L), median (IQR)   & 4.7 [2.4, 9.5] & 3.8 [2.4, 6.0] & 4.3 [2.0, 8.6] & 5.4 [2.3, 11.3] & 8.0 [3.9, 15.4] \\
Administration of a vasopressor, No. (\%) \\
\hspace{1em}Dopamine                         & 948 (4.8) & 79 (1.7) & 213 (3.4) & 387 (6.2) & 269 (9.9) \\
\hspace{1em}Phenylephrine                    & 6303 (31.8) & 2492 (53.9) & 1465 (23.5) & 1418 (22.7) & 928 (34.2) \\
\hspace{1em}Norepinephrine                   & 7450 (37.6) & 707 (15.3) & 1967 (31.5) & 2916 (46.7) & 1860 (68.5) \\
\hspace{1em}Vasopressin                      & 2475 (12.5) & 176 (3.8) & 455 (7.3) & 901 (14.4) & 943 (34.7) \\
\hspace{1em}Epinephrine                      & 1553 (7.8) & 542 (11.7) & 365 (5.8) & 289 (4.6) & 357 (13.2) \\
\hspace{1em}Dobutamine                       & 525 (2.6) & 18 (0.4) & 86 (1.4) & 219 (3.5) & 202 (7.4) \\
\bottomrule
\end{tabular}
}
\end{table*}

\begin{table*}[!htbp]
\centering
\begin{threeparttable}
\caption{The performance of NPCNet and benchmarks in the testing set}
\label{table2}
\begin{tabular}{>{\centering\arraybackslash}p{3cm}|cccc}
    \toprule
    Model &
    SI $\boldsymbol{\uparrow}$ &
    CHI $\boldsymbol{\uparrow}$ &
    DBI $\boldsymbol{\downarrow}$ &
    TDI $\boldsymbol{\uparrow}$ \\
    \midrule
    consensus K-means \cite{wilkerson2010consensusclusterplus} &
    -0.043 (0.011)*** &
    0.040 (0.005)*** &
    5.426 (0.870)*** &
    \textbf{0.716 (0.073)} \\

    KM-DTW &
    0.000 (0.000)*** &
    0.335 (0.004)*** &
    1.993 (0.011)*** &
    0.476 (0.093)*** \\

    DCN \cite{yang2017towards} &
    \underline{0.344 (0.048)}*** &
    \underline{1.482 (0.420)}** &
    \underline{1.140 (0.083)}*** &
    0.480 (0.080)*** \\

    DKM \cite{fard2020deep} &
    0.290 (0.053)*** &
    0.423 (0.029)*** &
    1.690 (0.098)*** &
    0.463 (0.065)*** \\

    USNID \cite{zhang2023clustering} &
    \textbf{0.711 (0.413)} &
    \textbf{4.415 (5.146)} &
    \textbf{0.830 (0.628)} &
    0.213 (0.197)*** \\

    naviDCN \cite{naviDCN} &
    0.335 (0.076)*** &
    0.824 (0.285)*** &
    1.184 (0.129)*** &
    \underline{0.533 (0.138)}*** \\
    \midrule
    NPCNet &
    \textbf{0.447 (0.012)} &
    \textbf{2.051 (0.161)} &
    \textbf{0.670 (0.022)} &
    \textbf{0.753 (0.061)} \\
    \bottomrule
\end{tabular}

\begin{tablenotes}[flushleft]
    \item The input was a tabular dataset for consensus K-means, DCN, and DKM. For static variables, we extracted the raw value directly from the EHRs. Missing values in static variables were imputed using mode for categorical variables and mean for continuous variables. For time-varying variables, we first implemented feature aggregation by selecting the most abnormal value for each variable. Missing values were then imputed using MICE \cite{van2011mice}. For KM-DTW, time-varying variables were aggregated into hourly values over the first six hours after ICU admission, resulting in a six-point time series. Pairwise distances were calculated using dynamic time warping (DTW) instead of the Euclidean distance used by K-means. We evaluated USNID using its semi-supervised version, based on the official implementation released by the authors \cite{zhang-etal-2021-textoir}. To ensure four clusters, two additional pseudo-labels were introduced for discharge status labels.

    \item ** $p<0.01$, *** $p<0.001$. For each model, we report the mean (SD) over 10 random seeds. Bold denotes the best mean score and scores not significantly different from it; underline denotes the best mean score among the remaining models.
\end{tablenotes}
\end{threeparttable}
\end{table*}

\subsection{Treatment Effects across Sepsis Phenotypes}
The median volume of cumulative intravenous (IV) fluids within 12 hours after initial hypotension was 2.49 liters (IQR, 1.33-4.14), and a vasopressor was initiated rapidly after initial hypotension, with more than 70\% of patients receiving the vasopressor in less than 1 hour.

IV fluid administration was comparable across phenotypes and only a few patients, especially those with the $\updelta$ phenotype, receive higher volumes. The time to vasopressor in the $\upalpha$ phenotype was earlier than in other phenotypes, while patients in the $\upbeta$ and $\upgamma$ phenotypes usually experienced a longer time to vasopressor (\textbf{Supplementary Fig. 6}).

\begin{figure*}[!htbp]
    \centering
    \includegraphics[width=\textwidth]{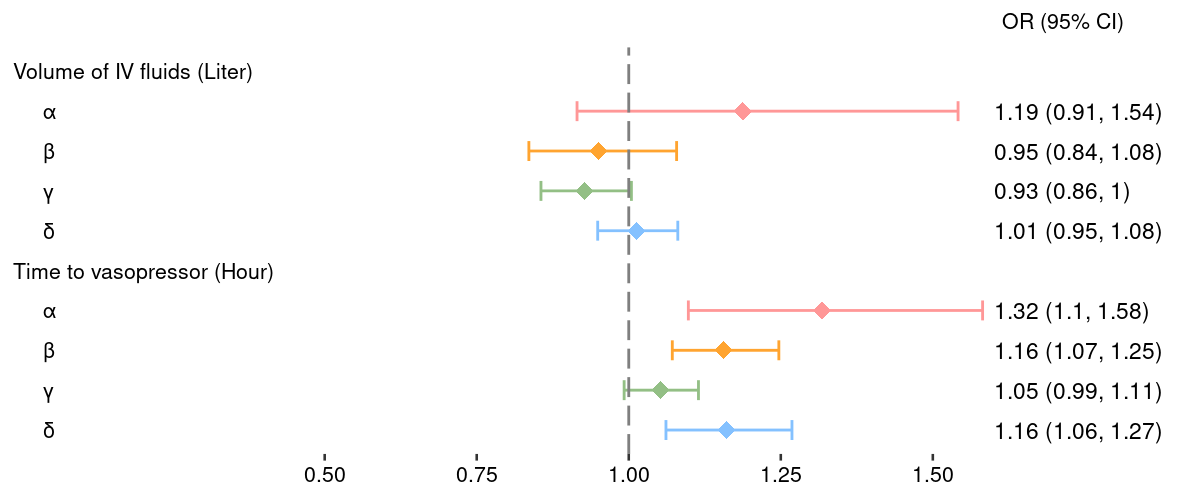}
    \caption{Multivariable logistic regression on in-hospital mortality with phenotypes.}
    \label{fig5}
\end{figure*}

The results of multivariable logistic regression in \textbf{Fig.} \ref{fig5} showed that, in all phenotypes, there was no statistically significant association between volume of IV fluids and in-hospital mortality.
However, the time to vasopressor exhibited phenotype-specific effects. 
A one-hour delay in the time to vasopressor initiation was associated with a 31.8\% increase in the odds of in-hospital mortality for patients with the $\alpha$ phenotype.
For the $\upbeta$ and $\updelta$ phenotypes, the corresponding increase was 15.6\% and 16.0\%, respectively, while no statistically significant association was observed for the $\upgamma$ phenotype.

\subsection{Internal Evaluation of Clustering Performance}
NPCNet achieved the best mean DBI and significantly outperformed most benchmarks (\textbf{Table} \ref{table2}). USNID obtained higher mean SI and CHI scores, but its large SDs across random seeds on all internal metrics indicate unstable performance, and none of its differences from NPCNet were significant. These results suggest that NPCNet produces cohesive, well-separated phenotypes with greater robustness to initialization.

\begin{figure*}[!htbp]
    \centering
    \includegraphics[width=\textwidth]{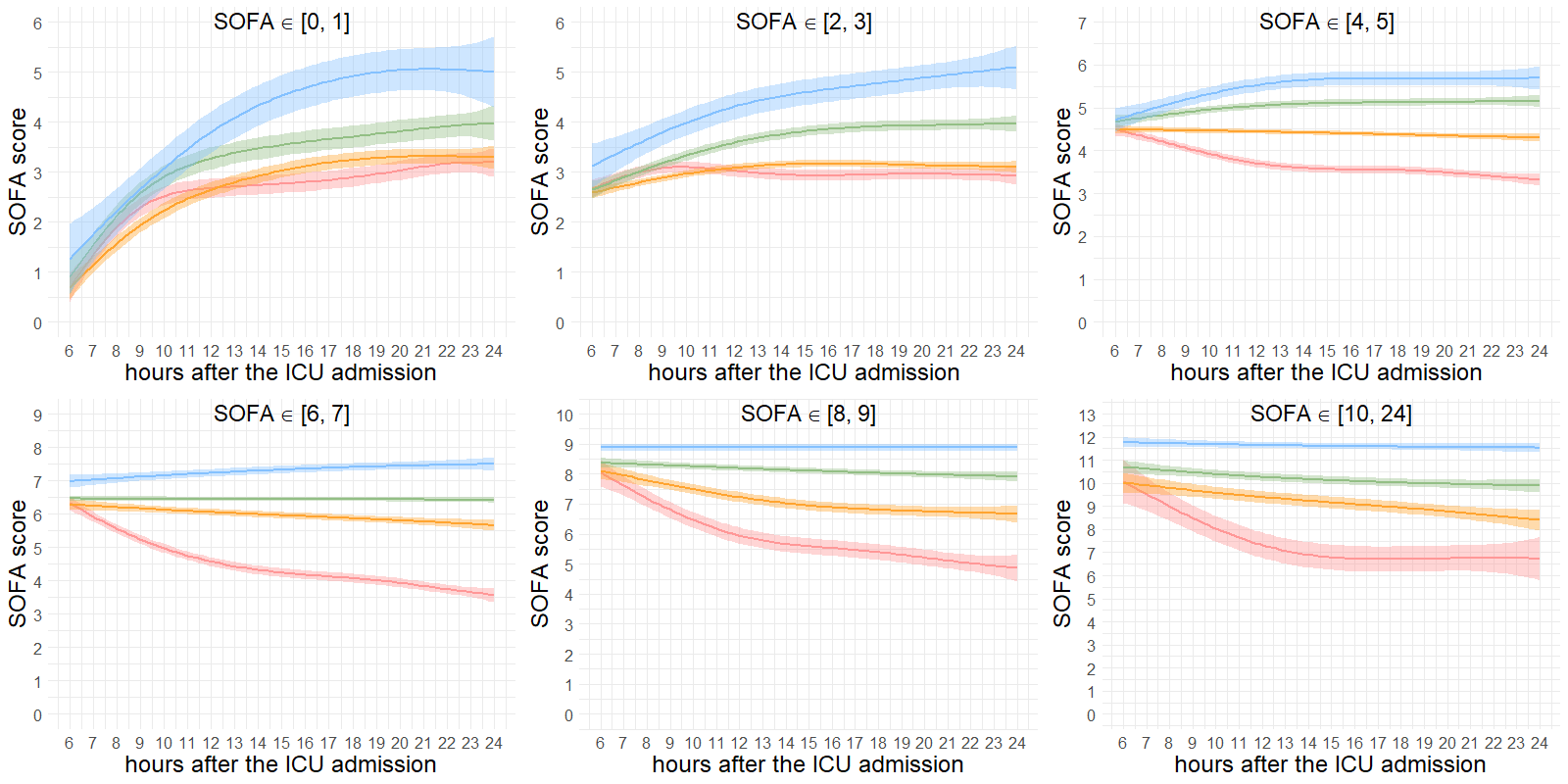}
    \caption{
    SOFA trajectories during the 18 hours following phenotype derivation by NPCNet, stratified by the SOFA score at six hours after ICU admission.}
    \label{fig6}
\end{figure*}

\begin{figure}[!htbp]
\centering
\subfloat[consensus K-means]{\includegraphics[width=0.235\linewidth]{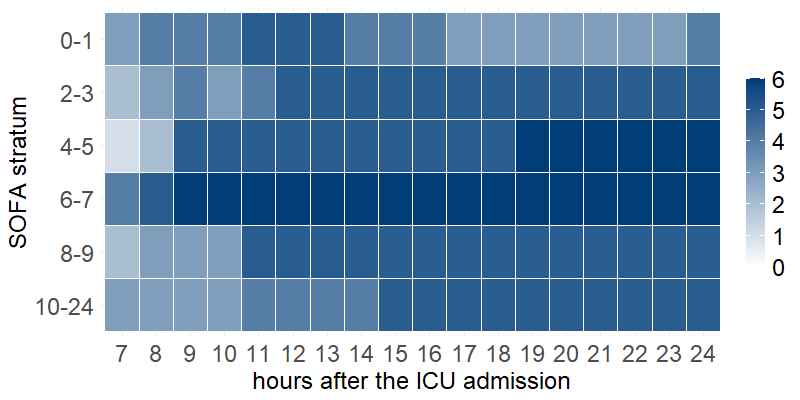}}
\subfloat[KM-DTW]{\includegraphics[width=0.235\linewidth]{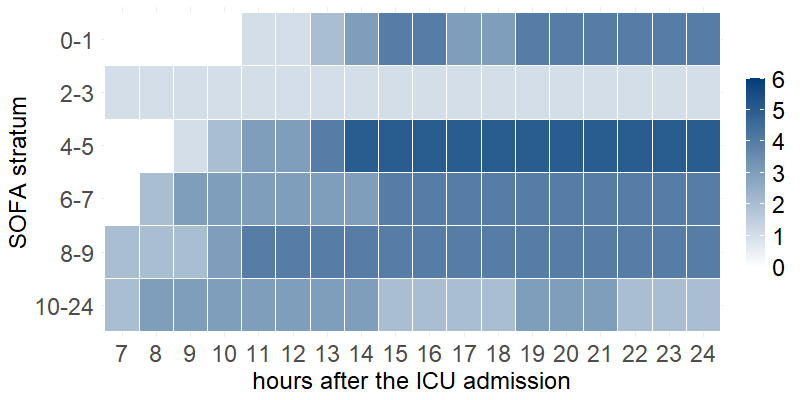}}
\subfloat[DCN]{\includegraphics[width=0.235\linewidth]{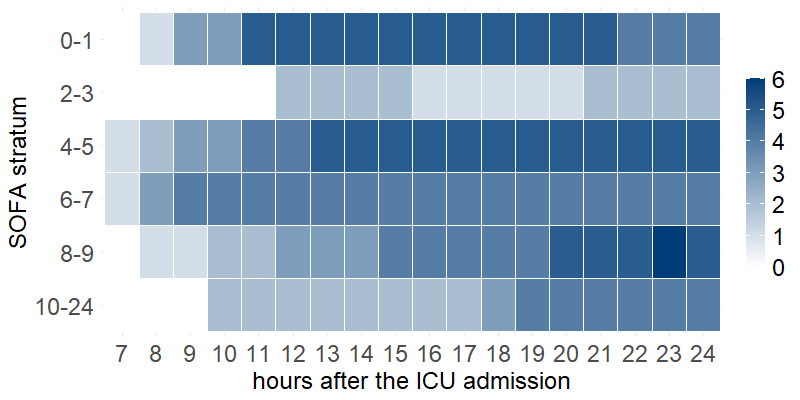}}
\subfloat[DKM]{\includegraphics[width=0.235\linewidth]{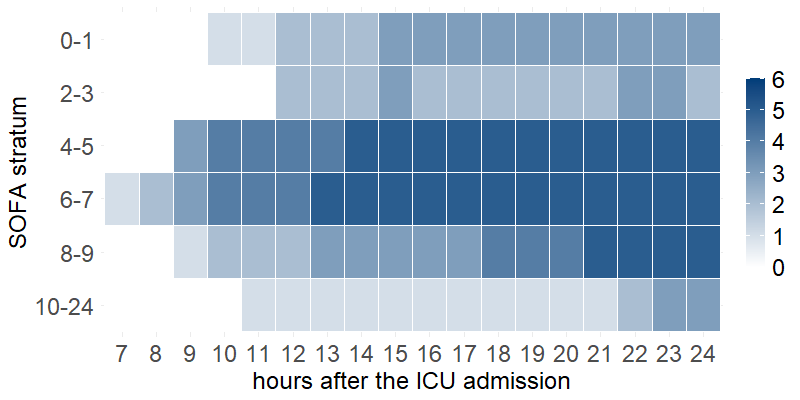}} \\
\subfloat[naviDCN]{\includegraphics[width=0.235\linewidth]{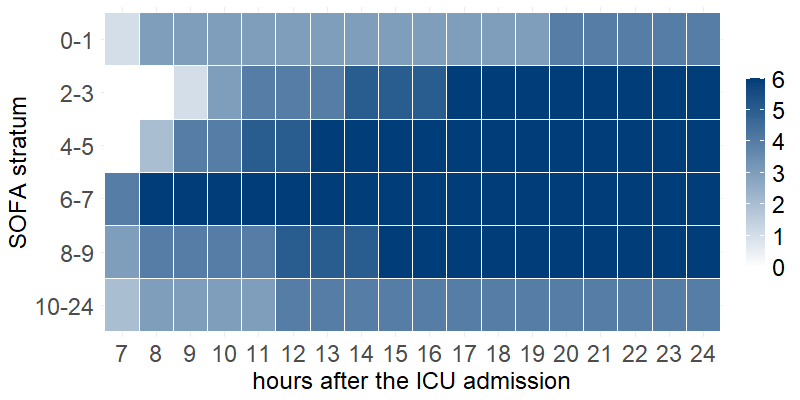}}
\subfloat[USNID]{\includegraphics[width=0.235\linewidth]{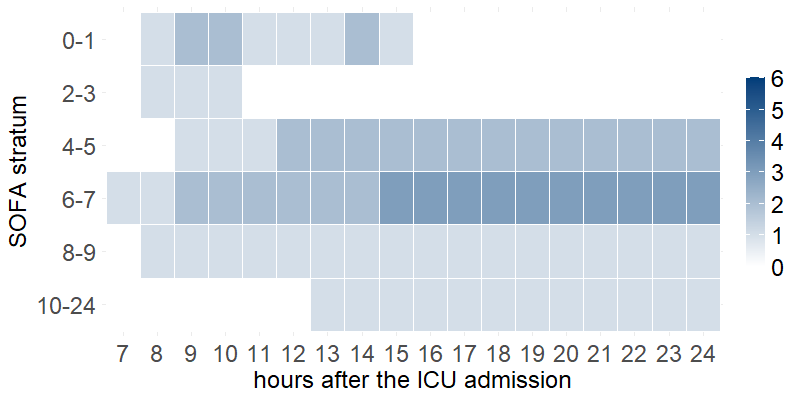}}
\subfloat[NPCNet]{\includegraphics[width=0.235\linewidth]{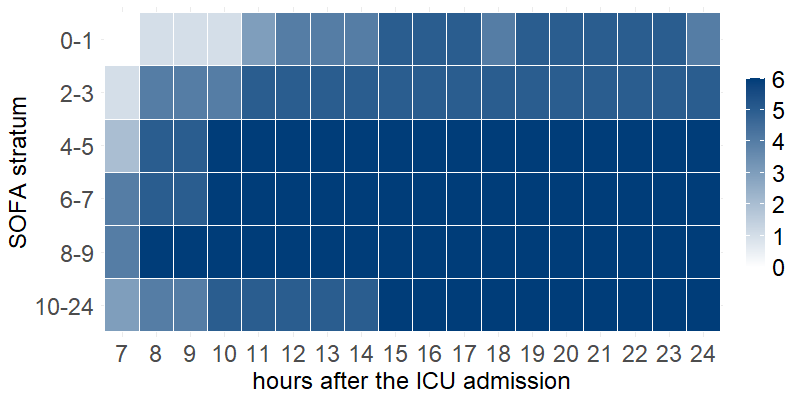}}
\caption{Pairwise comparisons of SOFA trajectories between four phenotypes at each hour within each stratum for NPCNet and benchmarks. The cell in the heatmap represents the number of significantly different phenotype pairs, out of six total possible pairwise combinations, at each time point within each SOFA stratum. The values range from 0 to 6, with darker colors indicating more pairwise differences, thereby reflecting how distinguishable the phenotypes are over time. The Trajectory Divergence Index (TDI) was then derived by normalizing the number of statistically significant pairs by the total number of pairwise comparisons, yielding a metric between 0 and 1 that quantifies the overall performance of models in clinical significance.}
\label{fig7}
\end{figure}

\subsection{Clinical Trajectories across Sepsis Phenotypes}
The four phenotypes exhibited distinct SOFA trajectories across nearly all stratifications, with disease severity generally increasing from the $\upalpha$ to the $\updelta$ phenotype, indicating distinct short-term patterns of organ dysfunction progression (\textbf{Fig.} \ref{fig6}).
To quantify the separability of these trajectories across models, we visualized the trajectories within each stratum as heatmaps and quantified their separability using TDI (\textbf{Fig.} \ref{fig7}). NPCNet consistently achieved higher divergence in most stratifications by approximately 10 hours after ICU admission, whereas naviDCN \cite{naviDCN} required longer for the phenotypes to demonstrate the divergence. Other models didn't exhibit a consistent pattern in divergence over time.

\subsection{External Validation}
We evaluated the generalizability of the sepsis phenotypes identified by NPCNet in the MIMIC-IV dataset using the eICU dataset. Similar to the clustering results in the MIMIC-IV dataset, both $\upalpha$ and $\updelta$ phenotypes exhibited abnormalities across multiple organ systems (\textbf{Supplementary Table 7}, \textbf{Supplementary Table 8}, and \textbf{Supplementary Fig. 7}), particularly in the hepatic and hematologic systems. Consistent with the findings from MIMIC-IV, these phenotypes were associated with substantially different clinical outcomes (\textbf{Supplementary Table 7}). The $\updelta$ phenotype exhibited increasing SOFA trajectories over time, whereas the $\upalpha$ phenotype showed decreasing SOFA trajectories (\textbf{Supplementary Fig. 8}).
In the treatment effect analysis, neither the volume of IV fluids nor the time to vasopressor showed a statistically significant association with in-hospital mortality across phenotypes (\textbf{Supplementary Figs. 9 }and\textbf{ 10}).

\subsection{Ablation Studies}
\begin{figure}[!htbp]
\centering
\subfloat[No Navigator]{\includegraphics[width=0.48\linewidth]{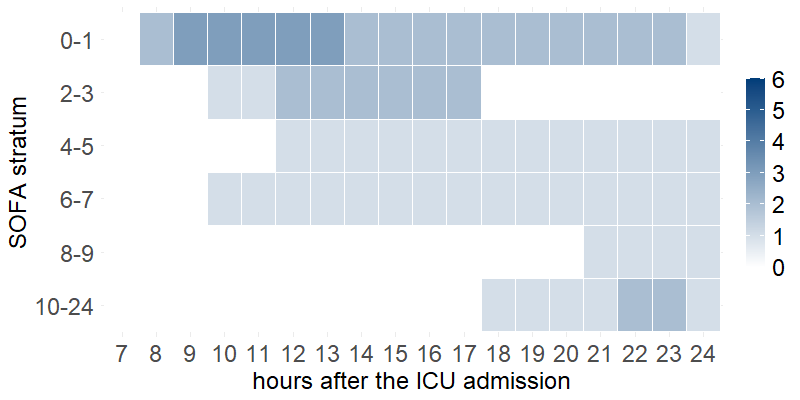}}
\hfill
\subfloat[ICU LOS Navigator]{\includegraphics[width=0.48\linewidth]{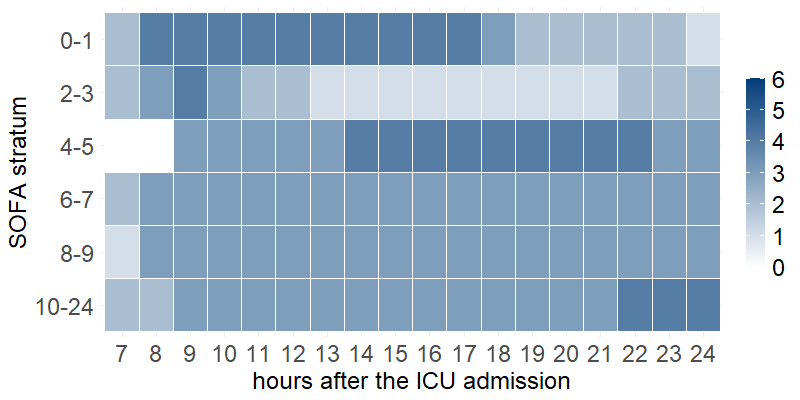}} \\
\subfloat[Discharge Status Navigator]{\includegraphics[width=0.48\linewidth]{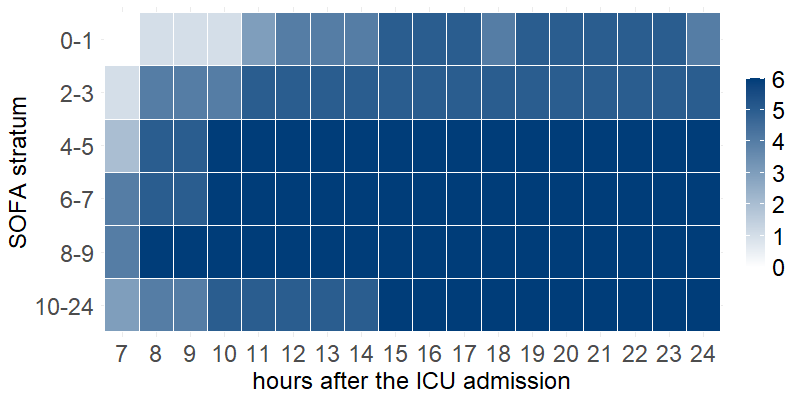}}
\hfill
\subfloat[Dual Navigator]{\includegraphics[width=0.48\linewidth]{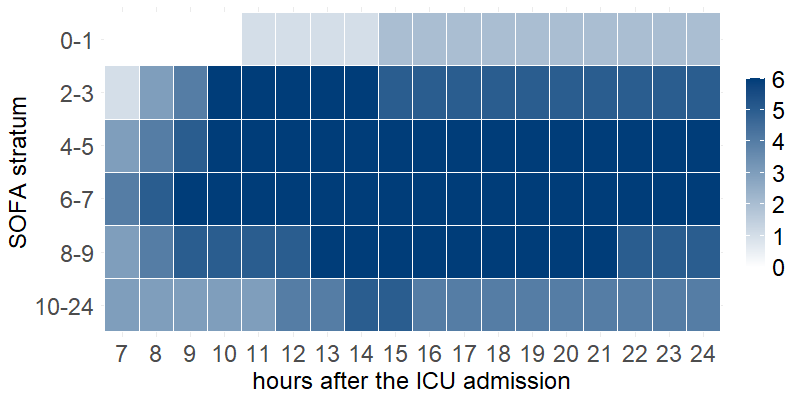}}
\caption{SOFA trajectories across phenotypes under different target navigators: No Navigator, ICU LOS Navigator, and Dual Navigator. In the Dual Navigator, ICU LOS guides $\mathcal{L}_{\text{prob}}$, while discharge status guides $\mathcal{L}_{\text{dist}}$. In the primary NPCNet, discharge status is used as the navigator. Pairwise comparisons of SOFA trajectories between phenotypes at each hour within each stratum in the testing set. The darker the color is, the greater the number of pairwise phenotypes that have statistically significant differences.}
\label{fig8}
\end{figure}

To evaluate the impact of each component, we conducted ablation studies on our proposed model. As shown in \textbf{Table} \ref{table3}, NPCNet achieved the best overall performance, whereas the table modalities consistently underperformed, and temporal matrices modalities showed competitive internal clustering but weaker clinical relevance. For text representation (\textbf{Table} \ref{table4}), binning improved TDI, while examination-order encoding improved internal clustering metrics; their combination achieved the best overall performance. Regarding objective functions (\textbf{Table} \ref{table5}), removing either navigator loss reduced clinical relevance, whereas jointly optimizing both achieved the best overall performance. As for target navigators, incorporating any navigator in the training stage substantially improved the differentiation of clinical outcomes compared to the model with no navigator (\textbf{Fig.} \ref{fig8}). In particular, NPCNet, which uses discharge status as the navigator, demonstrated the largest differences in in-hospital mortality across phenotypes (\textbf{Supplementary Fig. 11}). Moreover, despite not directly using ICU LOS as guidance, NPCNet exhibited competitive performance in the ICU LOS distribution (\textbf{Supplementary Fig. 12}). These findings suggested that the choice of target navigator plays a critical role in shaping clinical relevance. Additional ablation studies can be found in \textbf{Supplementary Table 9}.

\begin{table*}[!htbp]
\centering
\begin{threeparttable}
\caption{Ablation of the feature modalities and corresponding backbones in the testing set. The table uses the most abnormal measurements with imputation and an MLP backbone; temporal matrices aggregate measurements into hourly windows using ADL \cite{liu2023temporal} with a CNN backbone; pseudo text preserves the original temporal resolution without aggregation or imputation and uses a Transformer backbone (NPCNet).}
\label{table3}
\begin{tabular}{l|cccc}
    \toprule
    Feature modalities &
    SI $\boldsymbol{\uparrow}$ &
    CHI $\boldsymbol{\uparrow}$ &
    DBI $\boldsymbol{\downarrow}$ &
    TDI $\boldsymbol{\uparrow}$ \\
    \midrule
    table w/ MLP &
    0.134 (0.030)*** &
    0.298 (0.020)*** &
    2.180 (0.304)*** &
    \underline{0.595 (0.063)}*** \\

    matrix w/ CNN &
    \underline{0.428 (0.040)}** &
    \underline{1.441 (0.478)}** &
    \underline{0.766 (0.102)}* &
    0.587 (0.128)** \\
    \midrule
    NPCNet &
    \textbf{0.447 (0.012)} &
    \textbf{2.051 (0.161)} &
    \textbf{0.670 (0.022)} &
    \textbf{0.753 (0.061)} \\
    \bottomrule
\end{tabular}

\begin{tablenotes}[flushleft]
    \item * $p<0.05$, ** $p<0.01$, *** $p<0.001$. For each model, we report the mean (SD) over 10 random seeds. Bold denotes the best mean score and scores not significantly different from it; underline denotes the best mean score among the remaining models.
\end{tablenotes}
\end{threeparttable}
\end{table*}

\begin{table*}[!htbp]
\centering
\begin{threeparttable}
\caption{Ablation of the text representation by combining two value encodings (raw vs. discretized) and two positional encodings (timeline vs. examination order) in the testing set.}
\label{table4}
\begin{tabular}{>{\centering\arraybackslash}p{0.8cm}
                >{\centering\arraybackslash}p{2.2cm}|cccc}
    \toprule
    Text &
    Positional encoding &
    SI $\boldsymbol{\uparrow}$ &
    CHI $\boldsymbol{\uparrow}$ &
    DBI $\boldsymbol{\downarrow}$ &
    TDI $\boldsymbol{\uparrow}$ \\
    \midrule
    value & timeline &
    0.285 (0.051)*** &
    0.715 (0.364)*** &
    1.167 (0.167)*** &
    0.481 (0.072)*** \\

    value & order &
    \textbf{0.450 (0.027)} &
    \textbf{2.135 (0.164)} &
    \textbf{0.682 (0.024)} &
    0.576 (0.056)*** \\

    bin & timeline &
    \underline{0.348 (0.053)}*** &
    \underline{0.862 (0.245)}** &
    \underline{1.017 (0.083)}*** &
    \underline{0.681 (0.063)}* \\
    \midrule
    bin & order &
    \textbf{0.447 (0.012)} &
    \textbf{2.051 (0.161)} &
    \textbf{0.670 (0.022)} &
    \textbf{0.753 (0.061)} \\
    \bottomrule
\end{tabular}

\begin{tablenotes}[flushleft]
    \item ** $p<0.01$, *** $p<0.001$. For each model, we report the mean (SD) over 10 random seeds. Bold denotes the best mean score and scores not significantly different from it; underline denotes the best mean score among the remaining models.
\end{tablenotes}
\end{threeparttable}
\end{table*}

\begin{table*}[!htbp]
\centering
\begin{threeparttable}
\caption{Ablation of the objective functions in the testing set. Each navigator loss is removed individually while retaining reconstruction and clustering losses to prevent embedding collapse.}
\label{table5}
\begin{tabular}{cc|cccc}
    \toprule
    $\mathcal{L}_{\text{prob}}$ &
    $\mathcal{L}_{\text{dist}}$ &
    SI $\boldsymbol{\uparrow}$ &
    CHI $\boldsymbol{\uparrow}$ &
    DBI $\boldsymbol{\downarrow}$ &
    TDI $\boldsymbol{\uparrow}$ \\
    \midrule
    & &
    \underline{0.326 (0.012)}*** &
    \underline{0.817 (0.104)}*** &
    \underline{0.864 (0.017)}*** &
    0.268 (0.062)*** \\

    V & &
    0.278 (0.022)*** &
    0.512 (0.053)*** &
    1.262 (0.100)*** &
    0.600 (0.029)*** \\

    & V &
    0.211 (0.034)*** &
    0.319 (0.055)*** &
    1.910 (0.317)*** &
    \underline{0.633 (0.055)}*** \\

    \midrule
    V & V &
    \textbf{0.447 (0.012)} &
    \textbf{2.051 (0.161)} &
    \textbf{0.670 (0.022)} &
    \textbf{0.753 (0.061)} \\
    \bottomrule
\end{tabular}
\begin{tablenotes}[flushleft]
    \item ** $p<0.01$, *** $p<0.001$. For each model, we report the mean (SD) over 10 random seeds. Bold denotes the best mean score and scores not significantly different from it; underline denotes the best mean score among the remaining models.
\end{tablenotes}

\end{threeparttable}
\end{table*}

\section{Discussion}
Existing approaches to sepsis phenotyping face challenges in modeling temporal EHRs and incorporating clinical relevance into clustering. NPCNet addresses these challenges by preserving irregular temporal patterns while integrating clinical guidance into clustering. The resulting phenotypes showed distinct clinical characteristics, subsequent SOFA trajectories, and phenotype-specific treatment associations. Notably, SOFA trajectories were not used as training targets, providing an independent assessment of the clinical relevance of the identified phenotypes.

NPCNet is designed to balance three important objectives in sepsis phenotyping: generating representative patient embeddings, maintaining well-structured clusters in the embedding space, and aligning identified phenotypes with clinical significance. By integrating the target navigator into the deep clustering framework, NPCNet incorporates clinically relevant information into representation learning while maintaining the integrity of phenotypes in the embedding space. As shown in Table \ref{table2}, the results reveal a difference between internal clustering quality and clinical relevance. Although deep clustering methods such as DCN achieved competitive performance in internal metrics, their TDI scores were relatively modest. Conversely, consensus K-means achieved a relatively high TDI despite its poor internal clustering performance, suggesting that favorable mathematical clustering properties do not necessarily translate into clinically meaningful phenotypes. USNID achieved the highest mean scores in SI and CHI among the benchmarks, but exhibited substantial variability across random seeds and a relatively low TDI. This suggests that its strong internal clustering performance did not consistently translate into identifying phenotypes with distinct SOFA trajectories. In comparison, NPCNet achieved the best SI, DBI, and TDI scores and demonstrated competitive CHI performance. These results highlight the advantage of NPCNet in balancing internal clustering quality with clinical relevance.

Our experiments (Table \ref{table3}) showed that preserving the original temporal structure without conventional aggregation or imputation yielded better performance than the corresponding preprocessing alternatives. This may be attributed to the informative nature of measurement patterns in clinical practice. In EHRs, laboratory tests are ordered according to clinicians' assessments and suspected conditions, and their measurement frequency may therefore reflect underlying changes in patient status. Aggregating measurements or imputing missing values may obscure these patterns. Previous studies have similarly highlighted the complementary information provided by laboratory results and measurement patterns \cite{pivovarov2014identifying}. These findings support the importance of preserving both clinical values and their temporal information when modeling early sepsis phenotypes.

Moreover, the representation of numerical clinical measurements is an important consideration when applying language models to EHRs. Previous studies have reported limitations in the ability of language models to interpret numerical magnitudes, particularly when numerical values are tokenized into arbitrary subword units \cite{lin2020birds,zhang2020language,wallace2019nlp}. To address this issue, NPCNet discretizes continuous clinical measurements into bins before constructing pseudo-text representations. This design is supported by previous studies showing that bin-based representations can provide useful inductive biases for numerical learning and improve downstream performance in tabular tasks \cite{10.5555/3692070.3693143,gorishniy2022embeddings}. Consistent with these findings, our experiments (Table \ref{table4}) demonstrated that binning-based representations significantly outperformed representations using raw numerical values, suggesting that discretization facilitates more effective learning of clinical measurements within the proposed framework.

The ablation studies further demonstrate the contribution of the target navigator to clinical relevance. As shown in Table \ref{table5}, incorporating either navigator-related loss improved the clinical distinction between phenotypes, as reflected by their correspondence with clinical outcomes. Moreover, jointly optimizing the two navigator objectives, $\mathcal{L}{\text{prob}}$ and $\mathcal{L}{\text{dist}}$, achieved the best overall performance, suggesting that the two forms of clinical guidance provide complementary information. 

The choice of navigator target also influenced the clinical characteristics of the resulting phenotypes. To prevent data leakage, the target navigator was only trained using the training set, while the testing set was reserved exclusively for evaluation. In our experiments, the discharge status navigator produced phenotypes with substantial differences in in-hospital mortality while also demonstrating clinical distinctions in other outcomes, including ICU LOS. In contrast, using ICU LOS as the navigator yielded more limited improvements in outcomes beyond ICU LOS distribution. One possible explanation is the ambiguity of ICU LOS as a clinical target, as shorter stays may indicate either rapid recovery or early death. Consequently, patients with substantially different clinical conditions may receive similar LOS-related guidance, potentially limiting the navigator's ability to distinguish clinically distinct phenotypes. These findings suggest that discharge status may provide a more informative navigator for capturing overall clinical severity in the current framework. Nevertheless, navigator selection should ultimately depend on the clinical objectives of the phenotyping task.

Some of the identified phenotypes were similar to previous sepsis phenotyping studies \cite{seymour2019derivation}. Our $\upbeta$ phenotype showed relatively fewer abnormalities and patterns of organ dysfunction resembling the lower-risk $\upalpha$ phenotype, whereas $\updelta$ exhibited persistent multi-organ dysfunction and poor outcomes, similar to their high-risk $\updelta$ phenotype. NPCNet further extends these findings by capturing temporal trajectories using only the information in the first six hours of ICU admission, compared with approaches requiring longer observation periods, such as 72-hour SOFA trajectories \cite{xu2022sepsis}. Notably, patients within the same initial SOFA stratum exhibited divergent SOFA trajectories over the subsequent 18 hours. In particular, although $\upalpha$ and $\updelta$ patients presented with similarly severe initial conditions, their subsequent trajectories differed substantially, with $\upalpha$ showing a tendency toward improvement and $\updelta$ exhibiting persistent or progressive organ dysfunction. These findings suggest that NPCNet can identify differences in early disease progression that may not be apparent from initial severity assessments alone.

Treatment analyses also suggested phenotype-specific associations. IV fluid volume was not significantly associated with in-hospital mortality in any phenotype, consistent with previous findings on fluid timing and resuscitation strategies \cite{scheeren2019current}\cite{hidalgo2020delayed}\cite{shi2020vasopressors}. In contrast, earlier vasopressor administration was associated with lower mortality in the $\upalpha$, $\upbeta$, and $\updelta$ phenotypes, consistent with previous evidence supporting timely vasopressor initiation \cite{scheeren2019current,hidalgo2020delayed,shi2020vasopressors}. These findings suggest that treatment-outcome associations may vary across phenotypes and warrant further prospective evaluation.

To evaluate generalizability, we applied the model and binning thresholds derived from MIMIC-IV directly to the eICU dataset without re-training. In this external validation setting, the identified phenotypes exhibited similar clinical characteristics, in-hospital mortality, and SOFA trajectory patterns to those observed in the MIMIC-IV dataset. In contrast, treatment-related findings on time to vasopressor were less consistent across datasets, likely reflecting the known limitation in the eICU dataset, where incomplete documentation, such as drug infusion, has been previously mentioned \cite{pollard2018eicu}. Overall, these findings support the potential generalizability of NPCNet for identifying clinically meaningful sepsis phenotypes, while underscoring the need for caution when interpreting treatment-related findings across heterogeneous EHR datasets.

Our study has several limitations. First, NPCNet performs clustering at a single time point (6 hours after ICU admission) and does not capture potential phenotype transitions over time. Dynamic clustering with real-time tracking is an important direction for future research. Second, TDI evaluates whether SOFA trajectories differ between phenotypes but does not quantify the magnitude of divergence, and its value may depend on the chosen significance level. Further refinement of TDI is therefore needed to provide a more comprehensive assessment of clinical relevance across clustering models. Third, the retrospective observational design precludes causal inference regarding treatment effects, and retrospective timestamps may introduce measurement bias. Nevertheless, the findings provide preliminary evidence that may inform the development of future prospective studies.

\section{Conclusions}
We propose NPCNet, a deep clustering framework integrating a text embedding generator and a target navigator to enhance the clustering process to uncover sepsis phenotypes. The text embedding generator enables the model to capture patients’ temporal information without aggregation or imputation. By integrating the target navigator into the model, NPCNet improves the alignment of sepsis phenotypes with clinical relevance. NPCNet outperformed most benchmarks across internal and clinical metrics. Ablation studies confirmed the contribution of each component, and external validation on the eICU dataset supported its generalizability.

\section*{CRediT authorship contribution statement}
\textbf{Pi-Ju Tsai:} Conceptualization, Methodology, Software, Validation, Formal analysis, Investigation, Data curation, Writing -- original draft, Visualization. 
\textbf{Charkkri Limbud:} Validation, Writing -- review \& editing. 
\textbf{Kuan-Fu Chen:} Writing -- review \& editing, Supervision, Funding acquisition. 
\textbf{Yi-Ju Tseng:} Conceptualization, Methodology, Resources, Writing -- review \& editing, Supervision, Project administration, Funding acquisition.

\section*{Ethics statement}
This study utilized publicly available, fully de-identified data involving no direct interaction with human participants or identifiable records. Consequently, institutional ethical review and informed consent were exempt according to relevant institutional and national guidelines.

\section*{Declaration of competing interest}
The authors have no competing interests to declare.

\section*{Declaration of generative AI and AI-assisted technologies in the manuscript preparation process}
During the preparation of this work, the authors used generative AI tool (ChatGPT) to assist in improving the clarity, readability, and overall organization of the manuscript. The use of the tool was strictly limited to language editing and presentation enhancement, with all scientific contributions, data analyses, and conclusions developed entirely by the authors. After using these tools, the authors reviewed and edited the content as needed and take full responsibility for the content of the published article.

\section*{Acknowledgments}
This study was supported by grants from the National Science and Technology Council, Taiwan (NSTC 115-2221-E-A49-100-MY3 and 114-2634-F-A49-006), the CGMH-NYCU Joint Research Program (CGMH-NYCU-114-CORPG2P0072), and Chang Gung Memorial Hospital (CMRPG2P0342). The funders had no role in the study design and procedures; data collection, management, analysis, and interpretation; manuscript preparation, review, and approval; or the decision to submit the manuscript for publication.

\appendix
\section{Supplementary data}
Supplementary data to this article can be found.

\section*{Data availability}
MIMIC-IV is available from https://physionet.org/content/mimiciv/2.2/. The eICU Collaborative Research Database is publicly available at https://physionet.org/content/eicu-crd/2.0/. The source code for the proposed NPCNet framework is available at https://github.com/DHLab-TSENG/NPCNet.



\bibliographystyle{elsarticle-num}
\bibliography{sn-bibliography}

@article{singer2016third,
  title={The third international consensus definitions for sepsis and septic shock (Sepsis-3)},
  author={Singer, Mervyn and Deutschman, Clifford S and Seymour, Christopher Warren and Shankar-Hari, Manu and Annane, Djillali and Bauer, Michael and Bellomo, Rinaldo and Bernard, Gordon R and Chiche, Jean-Daniel and Coopersmith, Craig M and others},
  journal={Jama},
  volume={315},
  number={8},
  pages={801--810},
  year={2016},
  publisher={American Medical Association}
}

@article{rudd2020global,
  title={Global, regional, and national sepsis incidence and mortality, 1990--2017: analysis for the Global Burden of Disease Study},
  author={Rudd, Kristina E and Johnson, Sarah Charlotte and Agesa, Kareha M and Shackelford, Katya Anne and Tsoi, Derrick and Kievlan, Daniel Rhodes and Colombara, Danny V and Ikuta, Kevin S and Kissoon, Niranjan and Finfer, Simon and others},
  journal={The Lancet},
  volume={395},
  number={10219},
  pages={200--211},
  year={2020},
  publisher={Elsevier}
}

@article{prescott2019understanding,
  title={Understanding and enhancing sepsis survivorship. Priorities for research and practice},
  author={Prescott, Hallie C and Iwashyna, Theodore J and Blackwood, Bronagh and Calandra, Thierry and Chlan, Linda L and Choong, Karen and Connolly, Bronwen and Dark, Paul and Ferrucci, Luigi and Finfer, Simon and others},
  journal={American journal of respiratory and critical care medicine},
  volume={200},
  number={8},
  pages={972--981},
  year={2019},
  publisher={American Thoracic Society}
}

@article{seymour2017time,
  title={Time to treatment and mortality during mandated emergency care for sepsis},
  author={Seymour, Christopher W and Gesten, Foster and Prescott, Hallie C and Friedrich, Marcus E and Iwashyna, Theodore J and Phillips, Gary S and Lemeshow, Stanley and Osborn, Tiffany and Terry, Kathleen M and Levy, Mitchell M},
  journal={New England Journal of Medicine},
  volume={376},
  number={23},
  pages={2235--2244},
  year={2017},
  publisher={Mass Medical Soc}
}

@article{prescott2017improving,
  title={Improving long-term outcomes after sepsis},
  author={Prescott, Hallie C and Costa, Deena K},
  journal={Critical care clinics},
  volume={34},
  number={1},
  pages={175},
  year={2017}
}

@article{evans2021surviving,
  title={Surviving sepsis campaign: international guidelines for management of sepsis and septic shock 2021},
  author={Evans, Laura and Rhodes, Andrew and Alhazzani, Waleed and Antonelli, Massimo and Coopersmith, Craig M and French, Craig and Machado, Fl{\'a}via R and Mcintyre, Lauralyn and Ostermann, Marlies and Prescott, Hallie C and others},
  journal={Critical care medicine},
  volume={49},
  number={11},
  pages={e1063--e1143},
  year={2021},
  publisher={LWW}
}

@article{giamarellos2024pathophysiology,
  title={The pathophysiology of sepsis and precision-medicine-based immunotherapy},
  author={Giamarellos-Bourboulis, Evangelos J and Aschenbrenner, Anna C and Bauer, Michael and Bock, Christoph and Calandra, Thierry and Gat-Viks, Irit and Kyriazopoulou, Evdoxia and Lupse, Mihaela and Monneret, Guillaume and Pickkers, Peter and others},
  journal={Nature immunology},
  volume={25},
  number={1},
  pages={19--28},
  year={2024},
  publisher={Nature Publishing Group US New York}
}

@article{seymour2019derivation,
  title={Derivation, validation, and potential treatment implications of novel clinical phenotypes for sepsis},
  author={Seymour, Christopher W and Kennedy, Jason N and Wang, Shu and Chang, Chung-Chou H and Elliott, Corrine F and Xu, Zhongying and Berry, Scott and Clermont, Gilles and Cooper, Gregory and Gomez, Hernando and others},
  journal={Jama},
  volume={321},
  number={20},
  pages={2003--2017},
  year={2019},
  publisher={American Medical Association}
}

@article{xu2022sepsis,
  title={Sepsis subphenotyping based on organ dysfunction trajectory},
  author={Xu, Zhenxing and Mao, Chengsheng and Su, Chang and Zhang, Hao and Siempos, Ilias and Torres, Lisa K and Pan, Di and Luo, Yuan and Schenck, Edward J and Wang, Fei},
  journal={Critical Care},
  volume={26},
  number={1},
  pages={197},
  year={2022},
  publisher={Springer}
}

@inproceedings{yin2020identifying,
  title={Identifying sepsis subphenotypes via time-aware multi-modal auto-encoder},
  author={Yin, Changchang and Liu, Ruoqi and Zhang, Dongdong and Zhang, Ping},
  booktitle={Proceedings of the 26th ACM SIGKDD international conference on knowledge discovery \& data mining},
  pages={862--872},
  year={2020}
}

@inproceedings{yang2017towards,
  title={Towards k-means-friendly spaces: Simultaneous deep learning and clustering},
  author={Yang, Bo and Fu, Xiao and Sidiropoulos, Nicholas D and Hong, Mingyi},
  booktitle={international conference on machine learning},
  pages={3861--3870},
  year={2017},
  organization={PMLR}
}

@article{guo2020evaluation,
  title={An evaluation of time series summary statistics as features for clinical prediction tasks},
  author={Guo, Chonghui and Lu, Menglin and Chen, Jingfeng},
  journal={BMC medical informatics and decision making},
  volume={20},
  pages={1--20},
  year={2020},
  publisher={Springer}
}

@article{xie2022deep,
  title={Deep learning for temporal data representation in electronic health records: A systematic review of challenges and methodologies},
  author={Xie, Feng and Yuan, Han and Ning, Yilin and Ong, Marcus Eng Hock and Feng, Mengling and Hsu, Wynne and Chakraborty, Bibhas and Liu, Nan},
  journal={Journal of biomedical informatics},
  volume={126},
  pages={103980},
  year={2022},
  publisher={Elsevier}
}

@article{groenwold2020informative,
  title={Informative missingness in electronic health record systems: the curse of knowing},
  author={Groenwold, Rolf HH},
  journal={Diagnostic and prognostic research},
  volume={4},
  number={1},
  pages={8},
  year={2020},
  publisher={Springer}
}

@inproceedings{liu2023temporal,
  title={Temporal Phenotype Matrix Engineering for Electronic Health Records--Enhancing Coronary Artery Disease Prediction},
  author={Liu, Kuan-Hui and Chiang, Cheng-Yu and Wang, Hsin-Yao and Tseng, Yi-Ju},
  booktitle={2023 IEEE EMBS International Conference on Biomedical and Health Informatics (BHI)},
  pages={1--4},
  year={2023},
  organization={IEEE}
}

@article{zbMATH03129892,
 author = {Steinhaus, Hugo},
 title = {Sur la division des corps mat{\'e}riels en parties},
 fjournal = {Bulletin de l'Acad{\'e}mie Polonaise des Sciences, Classe 3},
 journal = {Bull. Acad. Pol. Sci., Cl. III},
 issn = {0001-4095},
 volume = {4},
 pages = {801--804},
 year = {1957},
 language = {French},
 zbMATH = {3129892},
 Zbl = {0079.16403}
}

@inproceedings{xie2016unsupervised,
  title={Unsupervised deep embedding for clustering analysis},
  author={Xie, Junyuan and Girshick, Ross and Farhadi, Ali},
  booktitle={International conference on machine learning},
  pages={478--487},
  year={2016},
  organization={PMLR}
}

@article{lin2020birds,
  title={Birds have four legs?! numersense: Probing numerical commonsense knowledge of pre-trained language models},
  author={Lin, Bill Yuchen and Lee, Seyeon and Khanna, Rahul and Ren, Xiang},
  journal={arXiv preprint arXiv:2005.00683},
  year={2020}
}

@inproceedings{10.5555/3692070.3693143,
author = {Lee, Kyungeun and Sim, Ye Seul and Cho, Hye-Seung and Eo, Moonjung and Yoon, Suhee and Yoon, Sanghyu and Lim, Woohyung},
title = {Binning as a pretext task: improving self-supervised learning in tabular domains},
year = {2024},
publisher = {JMLR.org},
booktitle = {Proceedings of the 41st International Conference on Machine Learning},
articleno = {1073},
numpages = {19},
location = {Vienna, Austria},
series = {ICML'24}
}

@inproceedings{hegselmann2023tabllm,
  title={Tabllm: Few-shot classification of tabular data with large language models},
  author={Hegselmann, Stefan and Buendia, Alejandro and Lang, Hunter and Agrawal, Monica and Jiang, Xiaoyi and Sontag, David},
  booktitle={International Conference on Artificial Intelligence and Statistics},
  pages={5549--5581},
  year={2023},
  organization={PMLR}
}

@inproceedings{rupp2023exbehrt,
  title={Exbehrt: Extended transformer for electronic health records},
  author={Rupp, Maurice and Peter, Oriane and Pattipaka, Thirupathi},
  booktitle={International Workshop on Trustworthy Machine Learning for Healthcare},
  pages={73--84},
  year={2023},
  organization={Springer}
}

@article{vaswani2017attention,
  title={Attention is all you need},
  author={Vaswani, Ashish and Shazeer, Noam and Parmar, Niki and Uszkoreit, Jakob and Jones, Llion and Gomez, Aidan N and Kaiser, {\L}ukasz and Polosukhin, Illia},
  journal={Advances in neural information processing systems},
  volume={30},
  year={2017}
}

@article{ferreira2001serial,
  title={Serial evaluation of the SOFA score to predict outcome in critically ill patients},
  author={Ferreira, Flavio Lopes and Bota, Daliana Peres and Bross, Annette and M{\'e}lot, Christian and Vincent, Jean-Louis},
  journal={Jama},
  volume={286},
  number={14},
  pages={1754--1758},
  year={2001},
  publisher={American Medical Association}
}

@article{johnson2023mimic,
  title={MIMIC-IV, a freely accessible electronic health record dataset},
  author={Johnson, Alistair EW and Bulgarelli, Lucas and Shen, Lu and Gayles, Alvin and Shammout, Ayad and Horng, Steven and Pollard, Tom J and Hao, Sicheng and Moody, Benjamin and Gow, Brian and others},
  journal={Scientific data},
  volume={10},
  number={1},
  pages={1},
  year={2023},
  publisher={Nature Publishing Group UK London}
}

@article{shankar2016developing,
  title={Developing a new definition and assessing new clinical criteria for septic shock: for the third international consensus definitions for sepsis and septic shock (Sepsis-3)},
  author={Shankar-Hari, Manu and Phillips, Gary S and Levy, Mitchell L and Seymour, Christopher W and Liu, Vincent X and Deutschman, Clifford S and Angus, Derek C and Rubenfeld, Gordon D and Singer, Mervyn and others},
  journal={Jama},
  volume={315},
  number={8},
  pages={775--787},
  year={2016},
  publisher={American Medical Association}
}

@article{moor2023predicting,
  title={Predicting sepsis using deep learning across international sites: a retrospective development and validation study},
  author={Moor, Michael and Bennett, Nicolas and Ple{\v{c}}ko, Drago and Horn, Max and Rieck, Bastian and Meinshausen, Nicolai and B{\"u}hlmann, Peter and Borgwardt, Karsten},
  journal={EClinicalMedicine},
  volume={62},
  year={2023},
  publisher={Elsevier}
}

@article{wilkerson2010consensusclusterplus,
  title={ConsensusClusterPlus: a class discovery tool with confidence assessments and item tracking},
  author={Wilkerson, Matthew D and Hayes, D Neil},
  journal={Bioinformatics},
  volume={26},
  number={12},
  pages={1572--1573},
  year={2010},
  publisher={Oxford University Press}
}

@article{fard2020deep,
  title={Deep k-means: Jointly clustering with k-means and learning representations},
  author={Fard, Maziar Moradi and Thonet, Thibaut and Gaussier, Eric},
  journal={Pattern Recognition Letters},
  volume={138},
  pages={185--192},
  year={2020},
  publisher={Elsevier}
}

@inproceedings{naviDCN,
  title={naviDCN: Navigator-Guided Multi-Modal Deep Clustering for Sepsis Phenotyping in Early ICU Admission},
  author={Tsai, Pi-Ju and Chen, Kuan-Fu and Limbud, Charkkri and Tseng, Yi-Ju},
  booktitle={2025 47th Annual International Conference of the IEEE Engineering in Medicine and Biology Society (EMBC)},
  pages={1--7},
  year={2025},
  organization={IEEE}
}

@article{van2011mice,
  title={mice: Multivariate imputation by chained equations in R},
  author={Van Buuren, Stef and Groothuis-Oudshoorn, Karin},
  journal={Journal of statistical software},
  volume={45},
  pages={1--67},
  year={2011}
}

@article{zhang2018identification,
  title={Identification of subclasses of sepsis that showed different clinical outcomes and responses to amount of fluid resuscitation: a latent profile analysis},
  author={Zhang, Zhongheng and Zhang, Gensheng and Goyal, Hemant and Mo, Lei and Hong, Yucai},
  journal={Critical Care},
  volume={22},
  pages={1--11},
  year={2018},
  publisher={Springer}
}

@article{wallace2019nlp,
  title={Do NLP models know numbers? probing numeracy in embeddings},
  author={Wallace, Eric and Wang, Yizhong and Li, Sujian and Singh, Sameer and Gardner, Matt},
  journal={arXiv preprint arXiv:1909.07940},
  year={2019}
}

@article{zhang2020language,
  title={Do language embeddings capture scales?},
  author={Zhang, Xikun and Ramachandran, Deepak and Tenney, Ian and Elazar, Yanai and Roth, Dan},
  journal={arXiv preprint arXiv:2010.05345},
  year={2020}
}

@article{gorishniy2022embeddings,
  title={On embeddings for numerical features in tabular deep learning},
  author={Gorishniy, Yury and Rubachev, Ivan and Babenko, Artem},
  journal={Advances in Neural Information Processing Systems},
  volume={35},
  pages={24991--25004},
  year={2022}
}

@article{pivovarov2014identifying,
  title={Identifying and mitigating biases in EHR laboratory tests},
  author={Pivovarov, Rimma and Albers, David J and Sepulveda, Jorge L and Elhadad, No{\'e}mie},
  journal={Journal of biomedical informatics},
  volume={51},
  pages={24--34},
  year={2014},
  publisher={Elsevier}
}

@article{scheeren2019current,
  title={Current use of vasopressors in septic shock},
  author={Scheeren, Thomas WL and Bakker, Jan and De Backer, Daniel and Annane, Djillali and Asfar, Pierre and Boerma, E Christiaan and Cecconi, Maurizio and Dubin, Arnaldo and D{\"u}nser, Martin W and Duranteau, Jacques and others},
  journal={Annals of intensive care},
  volume={9},
  pages={1--12},
  year={2019},
  publisher={Springer}
}

@article{hidalgo2020delayed,
  title={Delayed vasopressor initiation is associated with increased mortality in patients with septic shock},
  author={Hidalgo, Daniel Colon and Patel, Jaimini and Masic, Dalila and Park, David and Rech, Megan A},
  journal={Journal of Critical Care},
  volume={55},
  pages={145--148},
  year={2020},
  publisher={Elsevier}
}

@article{shi2020vasopressors,
  title={Vasopressors in septic shock: which, when, and how much?},
  author={Shi, Rui and Hamzaoui, Olfa and De Vita, Nello and Monnet, Xavier and Teboul, Jean-Louis},
  journal={Annals of translational medicine},
  volume={8},
  number={12},
  pages={794},
  year={2020}
}

@article{vincent1996sofa,
  title={The SOFA (Sepsis-related Organ Failure Assessment) score to describe organ dysfunction/failure: On behalf of the Working Group on Sepsis-Related Problems of the European Society of Intensive Care Medicine (see contributors to the project in the appendix)},
  author={Vincent, J-L and Moreno, Rui and Takala, Jukka and Willatts, Sheila and De Mendon{\c{c}}a, Arnaldo and Bruining, Hajo and Reinhart, CK and Suter, PeterM and Thijs, Lambertius G},
  journal={Intensive care medicine},
  volume={22},
  number={7},
  pages={707--710},
  year={1996},
  publisher={Springer-Verlag Berlin/Heidelberg}
}

@article{gu2014circlize,
  title={" Circlize" implements and enhances circular visualization in R},
  author={Gu, Zuguang and Gu, Lei and Eils, Roland and Schlesner, Matthias and Brors, Benedikt},
  year={2014}
}

@article{pollard2018eicu,
  title={The eICU Collaborative Research Database, a freely available multi-center database for critical care research},
  author={Pollard, Tom J and Johnson, Alistair EW and Raffa, Jesse D and Celi, Leo A and Mark, Roger G and Badawi, Omar},
  journal={Scientific data},
  volume={5},
  number={1},
  pages={1--13},
  year={2018},
  publisher={Nature Publishing Group}
}

@article{zhang2023clustering,
  title={A clustering framework for unsupervised and semi-supervised new intent discovery},
  author={Zhang, Hanlei and Xu, Hua and Wang, Xin and Long, Fei and Gao, Kai},
  journal={IEEE Transactions on Knowledge and Data Engineering},
  volume={36},
  number={11},
  pages={5468--5481},
  year={2023},
  publisher={IEEE}
}

@inproceedings{zhang-etal-2021-textoir,
    title = "{TEXTOIR}: An Integrated and Visualized Platform for Text Open Intent Recognition",
    author = "Zhang, Hanlei  and Li, Xiaoteng  and Xu, Hua  and Zhang, Panpan and Zhao, Kang  and Gao, Kai",
    booktitle = "Proceedings of the 59th Annual Meeting of the Association for Computational Linguistics and the 11th International Joint Conference on Natural Language Processing: System Demonstrations",
    pages = "167--174",
    year = "2021",
    url = "https://aclanthology.org/2021.acl-demo.20",
    doi = "10.18653/v1/2021.acl-demo.20",
}
\end{document}